\documentclass[11pt]{article}

\usepackage[final]{acl}
\usepackage{amssymb}
\usepackage{multirow}
\usepackage{times}
\usepackage{latexsym}
\usepackage{amsmath}
\usepackage[T1]{fontenc}
\usepackage{float}

\usepackage[utf8]{inputenc}

\usepackage{microtype}

\usepackage{inconsolata}

\usepackage{graphicx}
\usepackage[capitalise]{cleveref}
\usepackage{url}
\usepackage{booktabs}
\usepackage{bbding}
\usepackage{makecell}
\usepackage{algorithm}
\usepackage{algpseudocode}
\usepackage{amssymb}
\usepackage{xcolor}
\usepackage{pifont} 
\usepackage{amsfonts} % for \mathfrak
\usepackage{mathrsfs} % for \mathscr
\usepackage{tabularx}
\usepackage{colortbl}
\usepackage{lscape}
\usepackage{wrapfig}
\usepackage{subfig} 
\definecolor{red}{rgb}{1.0, 0.0, 0.0}
\definecolor{green}{rgb}{0.0, 1.0, 0.0}

\newcommand{\headercolor}{\rowcolor{gray!15}}

\title{Towards Harmonized Uncertainty Estimation for Large Language Models}

\author{
    Rui Li$^{1}$,
    Jing Long$^{1}$,
    Muge Qi$^{1}$,
    Heming Xia$^{2}$,
    Lei Sha$^{3}$,
    Peiyi Wang$^{1}$,
    Zhifang Sui$^{1}$\thanks{~~Corresponding author} \\
        \normalsize{$^{1}$ Peking University}\quad
        \normalsize{$^{2}$ The Hong Kong Polytechnic University} \quad
        \normalsize{$^{3}$Beihang University}\\
    \texttt{o\_l1ru1@stu.pku.edu.cn}
}

\begin{document}
\maketitle
\begin{abstract}

To facilitate robust and trustworthy deployment of large language models (LLMs), it is essential to quantify the reliability of their generations through uncertainty estimation. While recent efforts have made significant advancements by leveraging the internal logic and linguistic features of LLMs to estimate uncertainty scores, our empirical analysis highlights the pitfalls of these methods to strike a harmonized estimation between \textit{indication}, \textit{balance}, and \textit{calibration}, which hinders their broader capability for accurate uncertainty estimation. To address this challenge, we propose CUE\raisebox{-0.52ex}{\includegraphics[width=0.0274\textwidth, keepaspectratio]{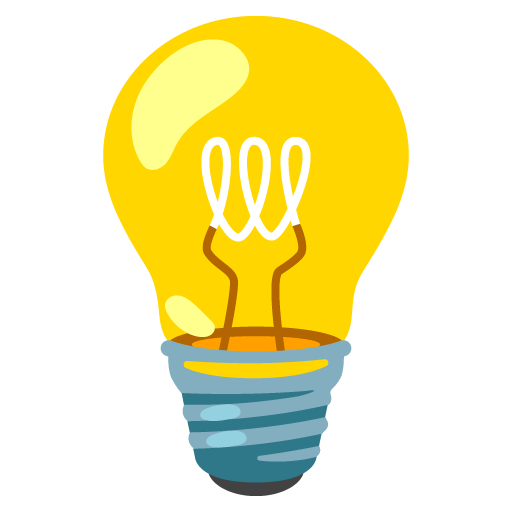}}(\underline{C}orrector for \underline{U}ncertainty \underline{E}stimation): A straightforward yet effective method that employs a lightweight model trained on data aligned with the target LLM's performance to adjust uncertainty scores. Comprehensive experiments across diverse models and tasks demonstrate its effectiveness, which achieves consistent improvements of up to $60\%$ over existing methods. Resources are available at \url{https://github.com/O-L1RU1/Corrector4UE}.

\end{abstract}

\section{Introduction}

% \fontspec{Symbola}\char"1F914}

% % \begin{quote}{
% \emoji{thinking-face}: \textit{``What do you have to give me confidence?''}\\
% \emoji{smiling-face}: \textit{``Uncertainty.''}
% % }\end{quote}  

\begin{quote}
    \raggedright \textit{Uncertainty is the only certainty there is.} \\
    \raggedleft - \textit{by John Allen Paulos}
\end{quote}

% \par\raggedleft--- By former superior
% and me(2024)

Large Language Models (LLMs) have demonstrated exceptional capabilities in handling a wide range of downstream tasks~\citep{openai:2023gpt4, Hugo:2023llama, Hugo:2023llama2, dubey2024llama}. They are gradually adopted as general-purpose API interfaces (e.g., ChatGPT\footnote{\url{https://chat.openai.com}}), providing valuable services and assistance in human life. Despite these impressive advancements, concerns persist regarding the tendency of LLMs to generate hallucinations and factual inaccuracies with confidence~\cite{zhang2023siren, wachter2024large}, which may mislead users to overestimate the reliability of the information provided by these models. To mitigate this issue, uncertainty estimation~\citep{loquercio2020general} proposed quantifying the reliability of model outputs so as to ensure the robustness and trustworthiness of AI-driven services. 

\textit{Harmonized} uncertainty estimation is expected to encompass three key aspects: \textbf{1) Indication}. The uncertainty score should clearly reflect the reliability of model responses, with higher scores signaling potential inaccuracies.  This can be framed as a classification task, with ``reliable'' or ``unreliable'' as the classes.  \textbf{2) Balance}. 
% When framing uncertainty estimation as a classification task, with uncertainty as the input and accuracy as the output, the task performance should achieve a balance between \textit{recall} and \textit{precision}, ensuring that challenging cases are appropriately flagged while minimizing the resources spent on false positives. 
Within classification framework, it's critical to strike a balance between recall and precision, ensuring that challenging cases are appropriately flagged while minimizing the resources spent on false positives. 
\textbf{3) Calibration}. The uncertainty score should align with human intuition and probabilistic expectations, to facilitate effective calibration.
% of the score and true situation. 
% if samples are assigned an uncertainty score of 0.8, there should be a 80\% chance that the model will answer incorrectly.
By striking a harmonized balance between these three aspects, uncertainty estimation provides an ideal measure of the model’s reliability, offering both usability and interpretability.

There has been growing interest in developing uncertainty estimation methods tailored for LLMs.
% , which can be broadly categorized into logit-based methods ~\citep{malinin2020uncertainty, kuhn2023semantic}, verbalized methods~\citep{lin2023generating, xiong2023can}, internal state-based methods~\citep{kadavath2022language, ji2024llm} and consistency-based methods~\citep{li2024think,
% pedapati2024large}.
% These meticulously designed training-free methods, which rely on the intrinsic logic and linguistic features of the LLMs, provide valuable insights into general uncertainty estimation.
However, with a thorough analysis across diverse uncertainty estimation methods, we found that there still remains a large performance gap between existing methods to achieve the harmonized uncertainty estimation. Specifically, methods that excel in one aspect fall short in others. For instance, SAR~\cite{duan2023shifting}, the outstanding and state-of-the-art method in the specific dataset SciQA~\cite{SciQA2023}, achieves the best performance in \textit{indication} but performs poorly in the view of \textit{calibration}. Furthermore, we found that the combination of uncertainty scores obtained by existing methods provides little improvement in uncertainty estimation performance, suggesting that these methods are quite homogeneous. These findings highlight considerable room for refinement in uncertainty estimation.

In this paper, we introduce CUE\raisebox{-0.4ex}{\includegraphics[width=0.028\textwidth, keepaspectratio]{picture/light-bulb_1f4a1.png}}, a simple yet effective framework for adjusting uncertainty scores, which is orthogonal to existing uncertainty estimation methods. 
% which can integrate smoothly with our framework.
% such as Semantic Entropy (SE)~\citep{kuhn2023semantic} and Shifting Attention to Relevance (SAR)~\citep{duan2023shifting}
% is correcting the inversion of rank ordering derived from model-dependent methods by integrating correction scores. 
% Accordingly, we propose an external insights-driven method to augment uncertainty estimation, orthogonal to existing advanced methods such as Semantic Entropy (SE)~\citep{kuhn2023semantic} and Shifting Attention to Relevance (SAR)~\citep{duan2023shifting}, both of which can be easily integrated with our method.
% transcending the inherent cognitive constraints of large models.
Specifically, we begin by curating dataset that is closely aligned with the target LLM's performance within a particular domain of knowledge.
This dataset is then utilized to train an auxiliary lightweight model, which serves as a \textit{Corrector} to adjust the uncertainty scores.
% This \textit{Corrector} is tasked with predicting the likelihood that the target LLM can accurately address specific questions.
By integrating the \textit{Corrector} trained on global alignment information with those uncertainty estimation methods that rely solely on the intrinsic logic and linguistic features of LLMs, we can significantly refine the uncertainty scores.

Our main contributions are thus as follows:

% We identify the limitations of current uncertainty estimation methods which xxxx

\begin{itemize}
    \item According to an empirical analysis of existing uncertainty estimation methods from both classification and calibration views, we found there is substantial room for improvement in their performance regarding classification indication, precision-recall balance, and calibration. 

    % The combination of different existing methods provides little to no improvement in uncertainty estimation performance. 
    % Notably, these methods cannot be effectively enhanced through weighted summation across different techniques, underscoring the potential for further refinement.
% suffer from inherent biases in LLMs, 
% including over-confidence and under-confidence. Additionally, we provide both theoretical proof and empirical evidence. 
    % specifically targeting logit-based methods.
    % (\cref{})
    \item We propose CUE\raisebox{-0.4ex}{\includegraphics[width=0.028\textwidth, keepaspectratio]{picture/light-bulb_1f4a1.png}}, an uncertainty score correction framework that employs a classifier, aligned with the model's task performance, as a \textit{Corrector} to adjust uncertainty scores.
    This \textit{Corrector} allows for seamless integration with existing uncertainty estimation methods. 
    % This approach can correct the inversion of uncertainty score rankings caused by xxxx.
    % the inherent biases of LLMs.
    % that refining the accuracy of uncertainty estimates by integrating globally informed uncertainties with those from conventional methods.
     
    % We propose a heuristic method that transcends the cognitive frameworks of LLMs by incorporating an observer’s perspective. This approach aims to enhance uncertainty estimation by providing a more nuanced understanding of the model’s capabilities, beyond what internal metrics alone can offer.
    \item  Extensive experiments and explorations are also conducted in areas such as generalization. The results demonstrate that our CUE\raisebox{-0.4ex}{\includegraphics[width=0.028\textwidth, keepaspectratio]{picture/light-bulb_1f4a1.png}} consistently enhances various existing uncertainty estimation methods, showing significant improvements in a harmonized manner across diverse data domains and target models.
    % in both relative and absolute terms. 
% Furthermore, we present comprehensive experimental evidence underscoring the robustness of our approach across diverse data domains and target models.

    % We demonstrate that our method achieves superior results across all orthogonal approaches. Additionally, we provide extensive experimental results showcasing its effectiveness and generalizability across different models, approaches, and datasets.

    % this approach significantly improves uncertainty estimation, both in relative and absolute terms.

\end{itemize}

\section{Related Work}
\label{section_related_work}

% \subsection{Relative Uncertainty and Absolute Uncertainty}
% \label{relative_and_absolute}
% Research on uncertainty estimation has led to two key concepts~\citep{kamath2020selective, vazhentsev2023hybrid}: \textit{relative uncertainty} and \textit{absolute uncertainty}, each providing distinct theories for interpreting and assessing the levels of uncertainty.
% Given an input $x$, a ground-truth answer $y$, and the predictive distribution of $Y$, the predictive uncertainty for the target model regarding the input $x$ is denoted as $\text{UE}(x, \theta)$.
% Relative uncertainty scores emphasize the accuracy of sample ranking, especially in indicating questions that the target model can correctly respond to from those it cannot. Ideally, for every pair $(x_i, y_i)$ and $(x_j, y_j)$ with their predictive distributions $Y_i$ and $Y_j$, we should have
% \begin{equation}    
% \begin{split}
% \text{UE}(x_i, \theta) &\leq \text{UE}(x_j, \theta) 
% \iff \\ P(Y_i = y_i | x_i,\theta) &\geq P(Y_j = y_j | x_j,\theta).
% \end{split}
% \end{equation}

% Stricter than relative uncertainty scores, absolute uncertainty scores support to represent the model's response accuracy.
% % In cases where there is 
% An 0.8 uncertainty score implies that the sample is expected to be answered correctly only 20\% of the time under similar conditions.
% This can be mathematically expressed as
% \begin{equation} 
% P(Y = y | \text{UE}(x, \theta) = q) = 1 - q. 
% \end{equation}

\subsection{Uncertainty Estimation for LLMs}
\begin{figure}[t]
    \centering
    \includegraphics[width=0.5\textwidth]{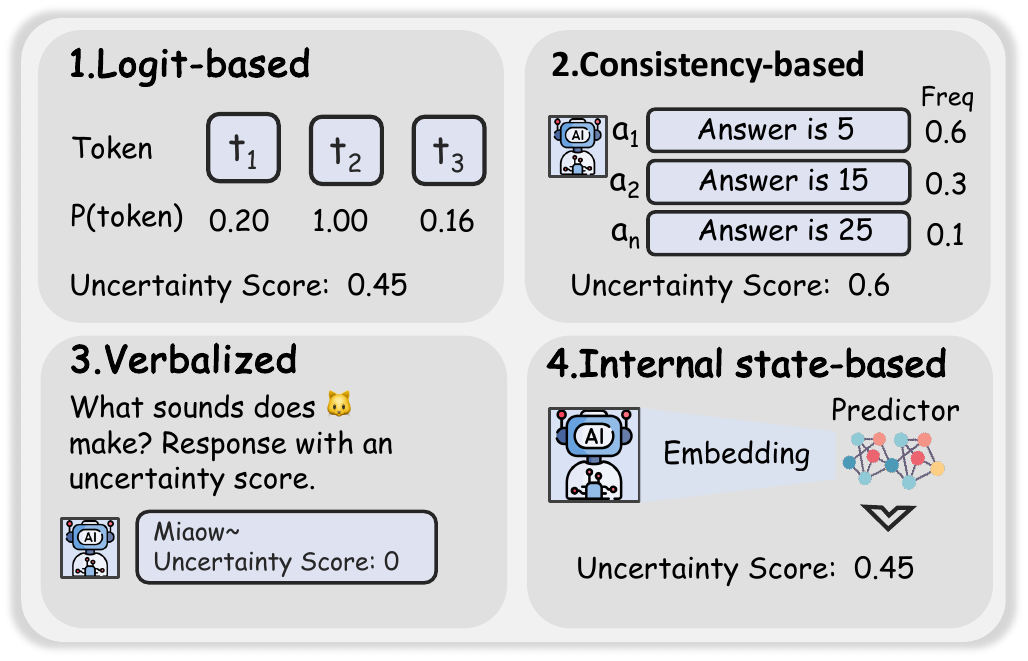}
    \caption{A concise overview figure of various uncertainty estimation method categories, including logit-based methods, verbalized methods, consistency-based methods, and internal state-based methods.}
    \label{fig:related_work}
\end{figure}

As illustrated in Figure \ref{fig:related_work}, uncertainty estimation methods for LLMs can be broadly categorized into logit-based methods, verbalized methods, consistency-based methods and internal state-based methods.

% \paragraph{Logit-based methods}
\textbf{Logit-based methods} are the most widely used and effective approaches in uncertainty estimation. 
% They derive uncertainty scores by converting model output logits using entropy-based techniques.
Predictive Entropy (PE)~\citep{malinin2020uncertainty} defined uncertainty as the entropy of the output logits distribution, which is widely adopted and built upon in subsequent research.
% serves as a foundational approach
% Predictive Entropy (PE)~\citep{malinin2020uncertainty}, the foundational logit-based uncertainty estimation method, defines total uncertainty as the entropy of the output distribution, which has been widely adopted by subsequent methods.
%Uncertainty estimation in autoregressive structured prediction
% After that, researchers proposed a series of methods based on the inherent characteristics of natural language generation to improve upon PE methods.
Following that, \citet{kuhn2023semantic} introduced semantic entropy (SE) that estimates uncertainty by marginalizing over semantically-equivalent samples in NLG tasks.
%Semantic uncertainty: Linguistic invariances for uncertainty estimation in natural language generation
% In the similar framework, \citet{nikitin2024kernel} employed positive semi-definite kernels and von Neumann entropy to capture semantic similarities. 
%Kernel Language Entropy: Fine-grained Uncertainty Quantification for LLMs from Semantic Similarities
% Furthermore, \citet{wang2024word} proposed Word-Sequence Entropy (WSE) to adjust uncertainty proportions at both the word and sequence levels based on semantic relevance, ensuring that uncertainty is aligned with the semantic importance of words within a response. 
%Word-Sequence Entropy: Towards Uncertainty Estimation in Free-Form Medical Question Answering Applications and Beyond
% In addition to measuring the similarity between generated responses, 
% \citet{wang2024word} proposed to judge the similarity between the target response and the generations.
%Semantic Density: Uncertainty Quantification in Semantic Space for Large Language Models
\citet{duan2023shifting} proposed Shifting Attention to Relevance (SAR), which focuses on relevant information and assigns significance weights to tokens based on their contributions to the overall response.
% \citet{duan2023shifting} proposed Shifting Attention to Relevance (SAR) to enhance predictive uncertainty quantification by focusing on relevant components at both token and sentence levels. 
%Shifting attention to relevance: Towards the uncertainty estimation of large language models
%MARS: Meaning-Aware Response Scoring for Uncertainty Estimation in Generative LLMs，bakman2024mars
% Unlike these carefully designed methods,
\citet{yaldiz2024not} introduced a Learnable Response Scoring Function (LARS), which utilizes supervised data to capture complex token-probability dependencies. 
%Do Not Design, Learn: A Trainable Scoring Function for Uncertainty Estimation in Generative LLMs
% While effective, the above methods are computationally expensive. 
% To alleviate these computational cost, \citet{kossen2024semantic} proposed Semantic Entropy Probes (SEPs) to approximate semantic entropy by leveraging hidden states from a single generation. 
%Semantic Entropy Probes: Robust and Cheap Hallucination Detection in LLMs
% Additionlcy, by combining Verbalized Uncertainty and Probing Uncertainty, \citet{tanneru2024quantifying} and \citet{liu2024examiningllmsuncertaintyexpression} both provide novel metrics for measuring model confidence which leverage direct model prompting and perturbation analysis, offering additional insights into the uncertainty associated with natural language explanations.
%Quantifying Uncertainty in Natural Language Explanations of Large Language Models
%Examining LLMs' Uncertainty Expression Towards Questions Outside Parametric Knowledge

\begin{figure*}[t]
    \centering
    \includegraphics[width=1\textwidth]{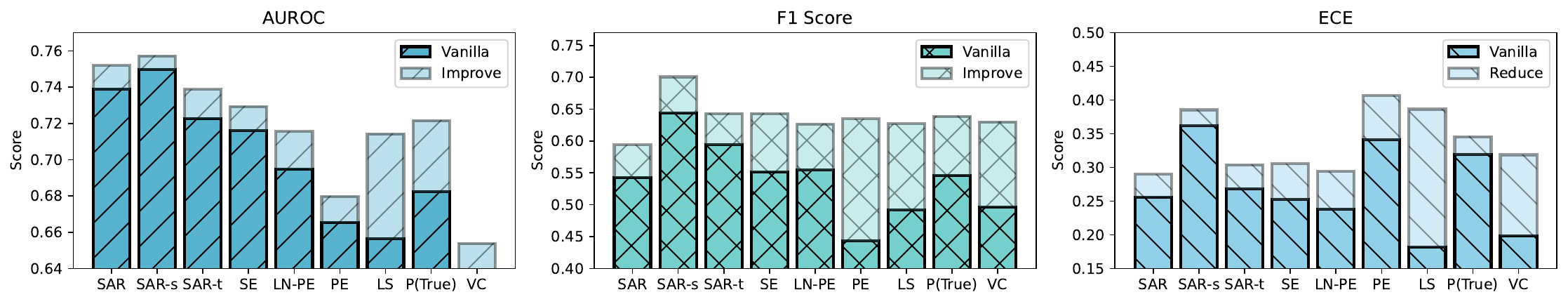}
    \caption{
    The performance of existing uncertainty estimation methods, evaluated on the SciQA dataset with the LLaMA-3-8B-Instruct model as the target, and the improvements after applying the \textit{Corrector}. Note that a lower ECE score indicates better performance, so we report its reduction.}
    \label{fig:preliminary_study}
\end{figure*}

\begin{figure}[h]
    \centering
    \includegraphics[width=0.35\textwidth]{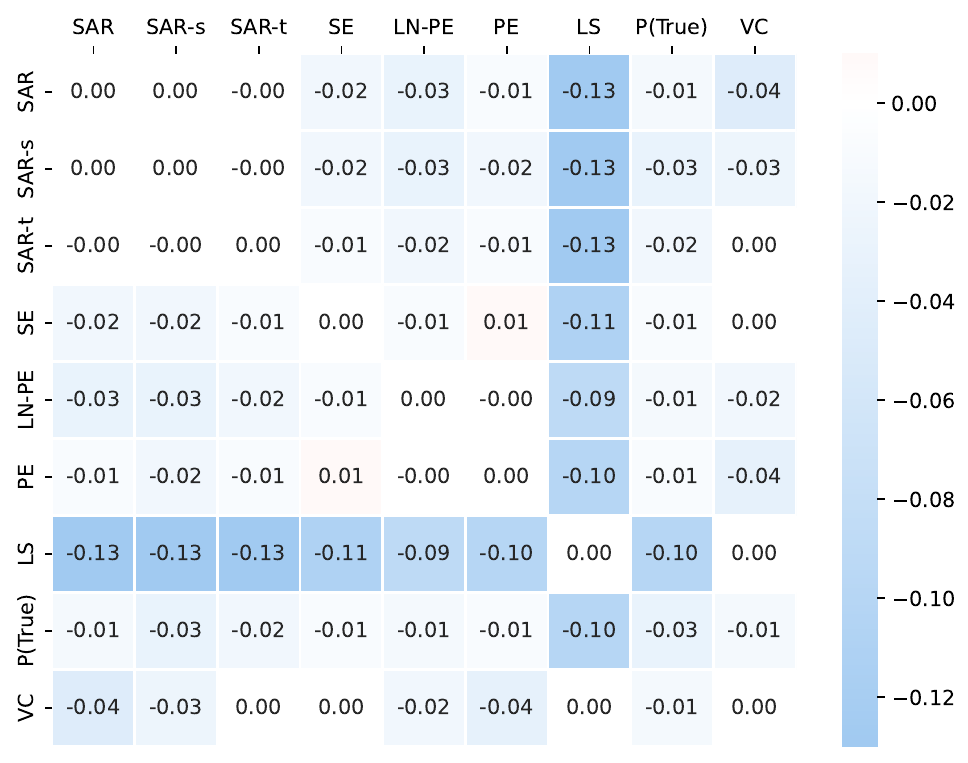}
    \caption{AUROC improvement across uncertainty scores combination from different existing methods.}
    \label{fig:comb}
\end{figure}

\textbf{Verbalized methods} \citep{xiong2023can,groot2024overconfidence} leverage LLMs' strong language and instruction-following abilities to express uncertainty, often by prompting the model to provide an uncertainty score. However, studies \citep{ni2024large,madhusudhan2024llms,becker2024cycles} have shown that LLMs struggle with faithfully conveying their uncertainties, particularly due to overconfidence. \textbf{Consistency-based methods}, such as those proposed by \citet{li2024think} and \citet{becker2024cycles}, assess uncertainty through multiple generated answers, using techniques like perturbation and aggregation to improve reliability. \citet{pedapati2024large} further reduced overconfidence by guiding LLMs to justify their answers. \textbf{Internal state-based methods} \citep{azaria2023internal,liu2024uncertainty} analyze LLM activations to predict errors, with \citet{kadavath2022language} and \citet{ji2024llm} exploring self-evaluation and probing estimators to enhance uncertainty estimation. 
% These methods, while promising, highlight the challenges of accurate uncertainty estimation in LLMs.

Due to space limitations, a more detailed discussion of related work is provided in the Appendix \ref{app:more_work}.

\section{Preliminary Study}
\begin{figure*}[t]
    \centering
    \includegraphics[width=1\textwidth]{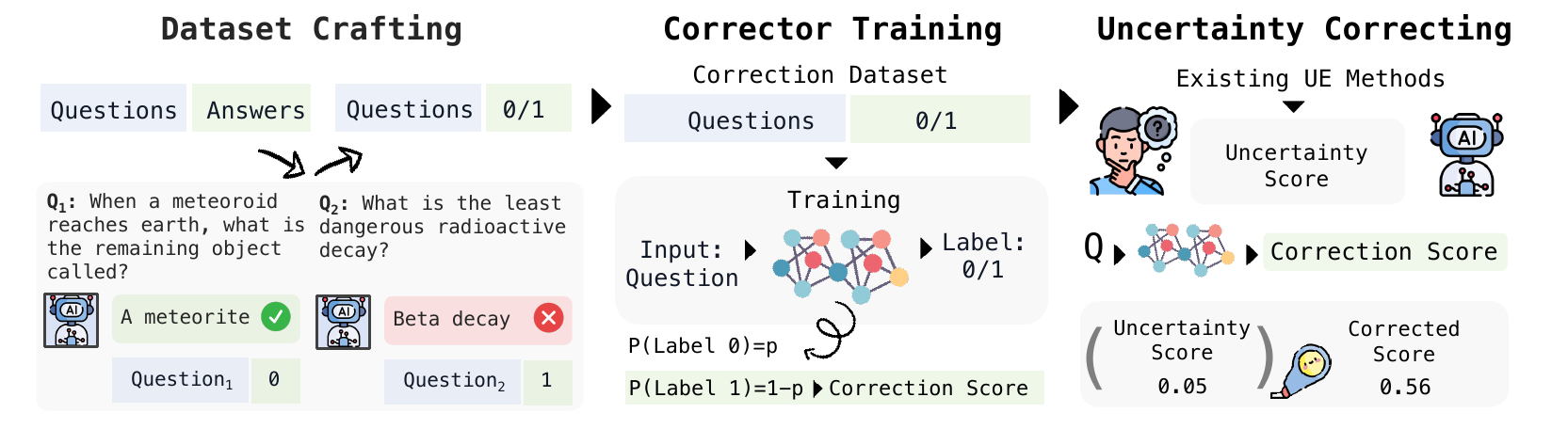}
    \caption{An overview of uncertainty score correction framework. 
Firstly, we construct a dataset that closely aligns with the target model's performance. This dataset is then utilized to train a lightweight auxiliary model that serves as a correction module, enabling seamless integration with existing uncertainty estimation methods to produce corrected uncertainty scores.}
    \label{fig:workflow}
\end{figure*}
\label{sec:preliminart_study}

\subsection{Limitation of Existing UE Methods}

We evaluate existing UE methods from both classification and calibration views, focusing on three key aspects of uncertainty scores: \textbf{indication, balance, and calibration}.
From the classification view, uncertainty scores are utilized to guide the classification process. Instances with scores above a threshold are classified as $c_1$ (unreliable) and those below as $c_0$ (reliable). 
We employ AUROC to measure how well the scores indicate unreliability and F1 score to evaluate their balance between precision and recall.
% We categorize instances into two classes: those the model can correctly response, labeled as $c_0$, and those it cannot, labeled as $c_1$.
% Instances with uncertainty scores above a threshold are classified as $c_1$, indicating a possible unreliable response, while those below are classified as $c_0$.
% Perfect uncertainty estimation should guarantee all instances belonging to class $c_1$ receive higher uncertainty scores than those belonging to $c_0$. 
% This  classification framework aligns with the concept of relative uncertainty we discussed in Section \ref{relative_and_absolute}.
% We employ AUROC to evaluate the performance of uncertainty estimation methods in producing \textbf{well-indicative} uncertainty scores and apply the F1 score to evaluate whether these uncertainty scores are \textbf{well-balanced} in precision and recall.
% The calibration view involves a more rigorous assessment and interpretation of uncertainty levels. These uncertainty scores are intended to align with human probabilistic intuition, offering more precise rankings of instances. To evaluate the model's calibration capability, we use ECE for \textbf{calibration} assessment.
The calibration view involves a more rigorous assessment and interpretation of uncertainty scores. Well-calibrated scores should align with human probabilistic intuition and provide more precise instance rankings. We use ECE to assess calibration.

\paragraph{Basic methods exhibit poor indication performance.}
Firstly, we focus on representative but naive methods including Lexical Similarity (LS)~\citep{fomicheva2020unsupervised}, Verbal Confidence (VC)~\citep{xiong2023can}, P(true)~\citep{kadavath2022language}, and Predictive Entropy (PE)~\citep{malinin2020uncertainty} that belong to four categories: consistency-based methods, verbal confidence methods, internal state-based methods, and logit-based methods, respectively.
% LS, VC, P(true), and PE 
% which are four each representative methods fall into distinct categories.
As shown in Figure \ref{fig:preliminary_study} and Table \ref{tab:main_res}, the AUROC scores for these methods across the target models and datasets exhibit general low performance, which is even close to random guessing. 
% These poor performances may stem from the complex of LLMs challenges naive implement. 
% especially in the TriviaQA dataset, with an average AUROC score of 0.48, 
% This also indicate a trade-off between ease of implementation and robustness for uncertainty estimation, revealing the significant optimization potential inherent in these methods.

\paragraph{Enhanced logit-based methods typically have low F1 scores.}

Some enhanced methods such as Length-normalized Predictive Entropy (LN-PE)~\citep{malinin2020uncertainty}, SAR-t, SAR-s, SAR, and Semantic Entropy (SE)~\citep{kuhn2023semantic}, 
% built on the foundational PE framework, 
make tailored adjustments 
% to tackle challenges observed in natural language generation tasks 
to refine predictive entropy process, which show improvements over PE in terms of AUROC.
% and demonstrate a promising classification performance.
% utilize the internal logic and language features of LLMs 
% These methods base their estimations on the target model's outputs—either at the logits or text level—and make tailored adjustments to tackle challenges observed in natural language generation tasks.
However, no one is universally optimal for all target models and datasets.
Moreover, as depicted in Figure \ref{fig:preliminary_study} and Figure \ref{fig:f1}, the F1 scores of those methods are particularly low.
This indicates that although those methods provided uncertainty scores with some potential to indicate the reliability of model response, they still fall short in striking a balance between precision and recall.
% , which means that there is a certain polarization tendency in the uncertainty scores, leading to an excessive number of false positives or false negatives.

\paragraph{Most existing methods fall short in calibration.}
As shown in Table \ref{tab:main_res} and Figure~\ref{fig:cal},  it appears that prior methods have overlooked the calibration aspect,  resulting in relatively poor performance in terms of ECE scores.

% Thus, from the perspective of classification, existing uncertainty estimation methods have considerable room for further improvement.

\subsection{Inter-method Cooperation}
\label{sec:inter-methed}

% We assess the complementarity of existing uncertainty estimation methods by evaluating the uncertainty scores obtained from pairwise combinations.

% 我们检查正交，

% To robustly prove the effectiveness of our \textit{Corrector} and differentiate it from existing uncertainty estimation methods, 
We examined whether the uncertainty scores derived from one uncertainty estimation method could refine the scores obtained from another method.
Specifically, we integrated the uncertainty scores from each method using the weighted combination and compared its performance with the top-performing method in the pair.
% To explore the potential complementarity between those different uncertainty estimation methods, 
As illustrated in Figure \ref{fig:comb}, these integrations do not enhance overall performance and may lead to a decline. 
This underscores the limitations in the complementary nature of existing methods.

% However, with a thorough analysis across diverse uncertainty estimation methods, we found that there still remains a large performance gap between existing methods to achieve the harmonized balance. Specifically, methods that excel in one aspect fall short in others. For instance, SAR~\cite{duan2023shifting}, the outstanding and state-of-the-art method in the specific dataset SciQA~\cite{SciQA2023}, achieves the best performance in \textit{indication} but performs poorly in the view of \textit{calibration}. Furthermore, we found that the combination of uncertainty scores obtained by existing methods provides little improvement in uncertainty estimation performance, suggesting that these methods are quite homogeneous. These findings highlight considerable room for refinement in uncertainty estimation.

Our analysis reveals a significant performance gap among existing methods in achieving harmonized uncertainty estimation, as individual methods excel in specific aspects but underperform in others. Furthermore, combining uncertainty scores from different methods yields minimal to no improvement, underscoring their homogeneous and non-complementary nature.

% , ablation study shows the effectiveness of the \textit{Corrector} in providing valuable heterogeneous gain to existing methods.

% The absence of complementarity among methods within the same category can be explained by their shared lineage. Enhanced methods often improve upon fundamental methods, leaving minimal room for additional gains when combined with fundamental methods. Furthermore, different enhanced methods fail to deliver performance improvements suggest a high degree of homogeneity in their optimization strategies.
% Cross-category methods also suffer from a lack of complementarity, likely due to their unstable performance and logical overlap. 

% This situation underscores the necessity of integrating external heterogeneous information, as facilitated by the \textit{Corrector}.

\section{Method}

\label{sec:method}

% Recognizing the inherent misalignment between model confidence and knowledge accuracy, coupled with the impracticality of modifying the models directly, 
In this section, we introduce CUE\raisebox{-0.4ex}{\includegraphics[width=0.028\textwidth, keepaspectratio]{picture/light-bulb_1f4a1.png}}, a correction framework featuring an intuitive approach to directly optimizing for uncertainty estimation, where a \textit{Corrector} is trained using a lightweight model to refine the uncertainty score.
Through this method, we provide a more robust solution for uncertainty estimation.
% In this section, we introduce an external insight-driven method to refine uncertainty estimation, 
% which integrates correction scores derived from a lightweight model trained on global information with those from existing uncertainty estimation approaches.
% Through this method, we provide a more robust solution for uncertainty estimation.
% effectively mitigating the adverse effects of inherent biases in LLMs.
% 这种方法利用 enables seamless integration with existing uncertainty estimation methods.
% This approach significantly enhances the accuracy of uncertainty scores across different models and datasets,
As shown in Figure \ref{fig:workflow}, Our method comprises three main steps including \textit{dataset crafting},\textit{ corrector training} and \textit{uncertainty correcting}. 
% This dataset is subsequently used to train a auxiliary lightweight model that functions as a correction module, facilitating seamless integration and enhancing performance across various methods.
% Firstly, we carefully construct a dataset that closely aligns with the target model's performance within a particular domain of knowledge.
% This dataset is then utilized to train an auxiliary lightweight model that serves as a correction module, facilitating seamless integration with existing uncertainty estimation methods to obtain corrected uncertainty scores.

% In this section, we introduce an external insights-driven method to refine uncertainty estimation.

% Our method examines uncertainty in meaning-space---the entropy of the random variable representing the output distribution in the semantic event-space.
% This is in contrast to entropy in the usual token event-space.
% To do this we introduce a novel algorithm for estimating the semantic equivalence relation as well as a novel uncertainty estimation algorithm for semantic entropy.
% At a high level this involves three steps:

\subsection{Dataset Crafting}
\label{s:data_construction}

% From the dataset $D$, extract data samples, consisting of a question $q_i$ and its corresponding ground truth answer $gt_i$. Input the question $q_i$ into the model $M$ to obtain the model-generated response $r_i$. Compare the model's response $r_i$ with the ground truth answer $gt_i$. If the response is correct, label it as 1; otherwise, label it as 0. This process creates a dataset that reflects, to some extent, the model's true cognitive ability.
% To construct a dataset that precisely aligns with the capabilities of the target model' knowledge,

We begin by extracting data from existing datasets to create an evaluation set for assessing the target model $M$'s performance in a specific domain. This set consists of a collection of question-answer pairs, denoted as $\mathcal{D} = \{(q_i, a_i) \mid i = 1, \ldots, n\}$. We then prompt $M$ to generate responses $r_i$ for each question $q_i$, forming a response set $\mathcal{R} = \{r_i \mid i = 1, \ldots, n\}$.
Subsequently, each response $r_i$ is subjected to a rigorous evaluation against the ground truth $a_i$, employing a hybrid approach that combines both rule-based and LLM-based methods. 
The rule-based method compares response $r_i$ to the ground truth $a_i$ using the longest common subsequence (LCS). 
A response $ r_i $ is considered equivalent to $a_i$ only if its ROUGE-L score, computed as $\text{ROUGE-L}( r_i,  a_i) = \frac{\text{LCS}( r_i,  a_i)}{\min(\text{len}( r_i), \text{len}( a_i))}$, is greater than threshold value, formalized as $\mathcal{M}_{Rule}(r_i , a_i) = \mathbb{I}_{\text{RougeL}(r_i, a_i) > 0.7}$. 
Additionally, we utilize GPT-turbo-3.5-0613~\citep{Ouyang0JAWMZASR22} to assess the equivalence between $ r_i $ and $a_i$ by directly prompting, formalized as $\mathcal{M}_{LLM}(r_i,a_i)=\mathbb{I}_{\text{True in LLM}(r_i, a_i)}$.
% To mitigate the risk of false negatives, we employ rigorous comparison strategies in both evaluations, such high threshold value and strict prompt rule.
% We adopted an ``OR'' computation approach for the judgment results derived from both rule-based and LLM-based methods by $
% \mathcal{M}(r_i, a_i) = \mathcal{M}_{\text{Rule}}(r_i, a_i) \lor \mathcal{M}_{\text{LLM}}(r_i, a_i)
% $,
% to prevent the omission of positive instances.

To mitigate false positives, we apply rigorous thresholds and strict prompting rules. The final judgment is determined using an ``OR'' logic: $\mathcal{M}(r_i, a_i) = \mathcal{M}_{\text{Rule}}(r_i, a_i) \lor \mathcal{M}_{\text{LLM}}(r_i, a_i)$, preventing the omission of positive instances.

After that, a binary label $c_i$ is assigned to each sample, defined as 
% $c_i=\mathcal{M}(r_i, a_i)$.
\begin{equation}
c_i \text{ = } \mathcal{M}(r_i, a_i)
\end{equation}
By pairing question $q_i$ with the label $c_i$, we form a correction dataset $\mathcal{D}_{\text{cor}} = \{(q_i, c_i) \mid i = 1, \ldots, n\}$, which serves as a representation of the target model's  performance in generating correct responses across a particular knowledge domain.
% external insight 
% into target model's performance in generating correct responses. 
To directly associate the questions with uncertainty, we transform the dataset form into $\mathcal{D^*}_{\text{cor}} = \{(q_i, 1-c_i) \mid i = 1, \ldots, n\}$. 

% we sample $M$ sequences $\{s^{(1)}, \dots, s^{(M)}\}$ which we will use later to estimate the uncertainty.
% These sequences must be sampled according to the distribution $p(\seq \mid x)$.
% In this paper, we sample these sequences only from a \textit{single} model using either multinomial sampling or multinomial beam sampling.
% We show in \cref{s:sampling_ablation}, that the choice of sampling temperature and sampling method can have a significant impact on the performance of both our method and the baselines.
% Unlike \citet{malinin2020uncertainty}, we do not use an ensemble of models.
% Ensembling would probably improve performance, but the cost of training multiple independent foundation models is often prohibitive.
\subsection{\textit{Corrector} Training}
\label{}

% Following the discussion in \ref{s:relative}, we frame uncertainty estimation as a classification task, focusing on the relative uncertainty score rankings between questions the model can answer correctly and those it cannot, thereby defining two distinct classes.

Employing the correction dataset $\mathcal{D^*}_{\text{cor}}$, we train a classifier to align with the performance of the target model. 
% The training objective is to determine whether the target model reliably responds to a given question.
Specifically, the classifier integrates a fully connected layer following a lightweight encoder model, such as RoBERTa~\citep{liu2019roberta} and Deberta~\citep{he2021debertav3}, with the representation of the special token $[CLS]$ as its input, denote as $\mathbf{h}_{[CLS]} \in \mathbb{R}^d$.
The output of the classifier is given by
$
\hat{y}_i = \sigma\left( \mathbf{W} \cdot \mathbf{h}_{[CLS]} + b \right),
$
where $\sigma(z)$ is the sigmoid function, used to compute the likelihood $y_i$ that a data point belongs to label $c_1$. During training, we minimize the binary cross-entropy loss function $\mathcal{L} = - \sum_{i=1}^{N} \left[ y_i \log(\hat{y}_i) + (1 - y_i) \log(1 - \hat{y}_i) \right]$ across the correction dataset.

% We obtain a \textit{Corrector}, an auxiliary component that can be integrated with existing uncertainty estimation methods to enhance the reliability of the process. Specifically, the \textit{Corrector} allows us to extract the probability that a sample belongs to category $c_1$ from its output. This probability can then be used to align the uncertainty scores with the performance of the target model, refining the uncertainty estimation process.

This results in a \textit{Corrector}, an auxiliary component that can be integrated with existing uncertainty estimation methods to enhance their reliability.

\subsection{Uncertainty Correcting}
% We utilize the \textit{Corrector} to improve the performance of uncertainty estimation through a complementary approach by 
% using the intrinsic logits and linguistic features of LLMs
% We integrate our \textit{Corrector} into uncertainty estimation frameworks to enhance their reliability. 
% The \textit{Corrector} allows us to extract the probability that a sample belongs to category $c_1$, as the correction score $C(x)$. This probability can then be used to align the uncertainty scores with the performance of the target model, refining the uncertainty estimation process.

% The \textit{Corrector} enables us to extract the probability that a instance belongs to category \(c_1\), represented as the correction score \(C(x)\), which is then used to align the uncertainty scores with the target model's performance, thereby refining the uncertainty estimation process.

We derive the probability that an instance \( x \) belongs to category \( c_1 \) from the \textit{Corrector}. This probability, denoted as the correction score \( C(x) \), can be utilized to adjust the uncertainty scores to align with the target model's performance, thereby refining the uncertainty estimation process.

In the refinement process, we first normalize the uncertainty scores generated from existing UE methods to match human probabilistic intuition, ensuring they fall within the range \([0, 1]\). Normalization is achieved via
$U_{\text{norm}}(x) = \frac{U(x) - \min(U)}{\max(U) - \min(U)}$, where \( U(x) \) represents the uncertainty score for a specific instance \( x \), computed by a chosen UE method. The terms \(\min(U)\) and \(\max(U)\) denote the minimum and maximum uncertainty scores across the entire dataset, respectively.
Following normalization, we apply our correcting by combining the normalized score \(U_{\text{norm}}(x)\) with the correction score \(C(x)\) generated by the \textit{Corrector}. 
The combination employs a weighted approach, where the corrected uncertainty score \(U_{\text{cor}}(x)\) is computed as:
\begin{equation}
U_{\text{cor}}(x) = w^* \cdot U_{\text{norm}}(x) + (1 - w^*) \cdot C(x)
\end{equation}
The optimal weight $w^*$ is determined through a grid search on the development dataset. 
This weighted method ensures that the corrected uncertainty scores balance the contributions of both the original and correction scores, thereby enhancing the reliability of the uncertainty estimation.

\section{Experiments}
\label{sec:exp}
% we demonstrate that our \textit{Corrector} is a effective robust module for enhancing the performance of uncertainty estimation in LLMs.

% We present comprehensive experimental results that demonstrate the effectiveness of our method across various models, datasets, and consistently surpassing all baselines.
\subsection{Experiments Setup}
\subsubsection{Models} 
% we adopt the OPT-2.7B~\citep{zhang2022opt} and LLaMA-3-8B-Instruct~\citep{dubey2024llama} as target models for our main experiments.
% Since model size is not the primary focus of our investigation, we select the 2.7B model from the OPT series, which is widely used in prior works~\citep{kuhn2023semantic,duan2023shifting}.
% And we choose the LLaMA-3-8B-Instruct for it
% To compare with previous studies~\citep{kuhn2023semantic,duan2023shifting}, 
% To compare with previous studies~\citep{kuhn2023semantic,duan2023shifting}, we adopt model from the OPT series~\citep{zhang2022opt}. 
% Since model size is not the primary focus of our investigation, 
\paragraph{Target models} We selected the OPT-6.7B\footnote{\url{huggingface.co/facebook/opt-6.7b}.}~\citep{zhang2022opt}, a model widely utilized in previous studies~\citep{kuhn2023semantic, duan2023shifting}, and the advanced open-source model LLaMA-3-8B-Instruct\footnote{\url{huggingface.co/meta-llama/Meta-Llama-3-8B-Instruct}.}~\citep{dubey2024llama} as the target models for our main experiments.

\paragraph{Base Models} We employed lightweight encoder models as the base model to train the \textit{Corrector}, including models from the RoBERTa series~\cite{liu2019roberta}  and DeBERTa series~\cite{he2021debertav3,he2021deberta}.
% , including RoBERTa-base\footnote{\url{huggingface.co/FacebookAI/roberta-base}.} and RoBERTa-large\footnote{\url{huggingface.co/FacebookAI/roberta-large}.}, 
% as well as models from the DeBERTa series~\cite{he2021debertav3,he2021deberta}, 
% such as DeBERTa-v3-large\footnote{\url{huggingface.co/microsoft/deberta-v3-large}.}, DeBERTa-v3-base\footnote{\url{huggingface.co/microsoft/deberta-v3-base}.}, DeBERTa-v3-small\footnote{\url{huggingface.co/FacebookAI/deberta-v3-small}.}, and DeBERTa-v3-xsmall\footnote{\url{huggingface.co/FacebookAI/deberta-v3-xsmall}.}. These models represent different types, series, and sizes. The results indicate that 

% Furthermore, to demonstrate the generalizability of our method, 

\subsubsection{Metrics}
\label{metric}
% \paragraph{AUROC} We use the area under the receiver operating characteristic curve (AUROC) as our primary metric for evaluating the performance of uncertainty estimation methods from a classification perspective. In our setting, an AUROC of 1 signifies that the uncertainty estimation method perfectly distinguishes between questions the target model can answer reliably and those it cannot, while an AUROC of 0.5 indicates that the estimation is no better than random guessing.
\paragraph{AUROC} We use the area under the receiver operating characteristic curve (AUROC) to evaluate uncertainty estimation methods from a classification view. In our setting, an AUROC of 1 signifies perfect indicative performance to distinguish between samples the target model can answer reliably and those it cannot, while an AUROC of 0.5 indicates that the estimation is no better than random guessing.
% while 0.5 represents random guessing.

% In our experiments, it can be used to evaluate the performance of \textit{relative uncertainty}.
% AUROC measures the methods' performance to rank questions target model cannot respond correctly with higher uncertainty than correct ones. 

\paragraph{F1 Score}
F1 score is used to evaluate the balance between precision and recall in classification tasks. It is the harmonic mean of precision and recall, where both are equally important. The F1 score ranges from 0 to 1, with 1 indicating perfect precision and recall.
% The primary goal of uncertainty estimation is to precisely pinpoint the "difficult" and "risky" instances in model predictions, while preventing the misclassification of non-risky points to minimize potential resource waste. The F1 score serves as an ideal metric for this purpose, as it considers both false positives and false negatives:
\begin{equation}
\text{F1 Score} = 2 \times \frac{\text{Precision} \times \text{Recall}}{\text{Precision} + \text{Recall}}
\end{equation}

\paragraph{ECE} We use Expected Calibration Error (ECE) to evaluate the performance of calibration, 
% A well-calibrated score is excepted to reflects the true likelihood of instances. 
% It can evaluate how well the predicted scores align with actual conditions, 
which is calculated by partitioning predicted confidence scores into bins and comparing the average confidence in each bin to the actual fraction of correct predictions, formalized as
\begin{equation}
\mathrm{ECE} = \sum_{m=1}^M \frac{|B_m|}{n} \left| \text{acc}(B_m) - \text{conf}(B_m) \right|
\end{equation}
% where $M$ is the number of bins, $B_m$ is the set of predictions in bin $m$, $|B_m|$ is the number of samples in that bin, and $n$ is the total number of samples. 
In the computing of ECE, we treat the confidence score as 1 minus uncertainty score. 

% Lower ECE values indicate predicted scores closely match observed outcomes, while higher ECE values suggest misalignment between predicted scores and actual performance. 

% The core objective of uncertainty estimation is to accurately identify the ``difficult'' and ``risky'' points in model predictions while avoiding the misclassification of non-risky points to reduce potential resource waste. The F1 score, provides an ideal metric for achieving this goal:  

% \begin{equation}
% \text{F1 Score} = 2 \times \frac{\text{Precision} \times \text{Recall}}{\text{Precision} + \text{Recall}},
% \end{equation}
% which accounts for both false positives and false negatives.

\subsubsection{Datasets}
% For evaluation purposes, 
We focus on the question-answering task using two representative datasets in the main experiments:
\textbf{TriviaQA}~\citep{joshi2017triviaqa}, and \textbf{SciQA}~\citep{SciQA2023}.
% and MedMCQA~\citep{pal2022medmcqa}.
% \paragraph{TriviaQA} 
TriviaQA comprises 95,000 question-answer pairs created by trivia enthusiasts, supplemented with independently sourced evidence documents. 
% We utilize TriviaQA as a closed-book task, where target models are challenged to provide answers without access to supporting paragraphs. 
% \paragraph{SciQA} 
% SciQA contains 2,565 question-answer pairs fetched from the open research knowledge graph, covering several research fields ranging from science and technology, life sciences to social sciences. 
SciQA contains 2,565 question-answer pairs fetched from the open research knowledge graph, covering several research fields ranging from science and technology like Computer Science, Engineering, Chemistry, and Geology, life sciences like Immunology and Genetics to social sciences like Economics and Urban Studies.

% \paragraph{MedMCQA} Beyond general QA tasks, our emphasis is on high-stakes decision-making, particularly in the medical domain. To this end, we employ MedMCQA, a large-scale multiple-choice question answering dataset designed for real-world medical entrance exams. To better align with practical applications, we convert the multiple-choice questions into a question-answer format, omitting the answer options.

\subsubsection{Baselines}
% We select various uncertainty estimation methods as baselines
% of techniques, particularly focusing on logits-based methods and their variants.

% and we divide them into two sets: 
We select a variety of representative uncertainty estimation methods as baselines, with a particular focus on logits-based methods. 

Among the baselines, we cover multiple categories, including logit-based, verbalized, internal state-based, and consistency-based methods, including: \textbf{Lexical Similarity (LS)}~\citep{fomicheva2020unsupervised}, which computes the similarity between multiple sentences as a measure of consistency; \textbf{Verbal Confidence (VC)}~\citep{xiong2023can}, which requires the target model to respond and provide a confidence score; \( \textbf{P(\text{True})} \)~\citep{kadavath2022language}, which first asks the target model to propose an answer and then evaluates it using an internal probability mechanism; and \textbf{Predictive Entropy (PE)}~\citep{malinin2020uncertainty}, which calculates uncertainty by measuring the entropy of the predictive posterior. 

We also explore a series of advanced logit-based methods including: \textbf{Length-normalized Predictive Entropy (LN-PE)}~\citep{malinin2020uncertainty}, which adjusts PE by normalizing it according to sentence length; \textbf{Semantic Entropy (SE)}~\citep{kuhn2023semantic}, which clusters sentences with equivalent meanings and calculating cluster-wise entropy; and \textbf{Shifting Attention to Relevance (SAR)}~\citep{duan2023shifting}, which encompasses \textbf{SAR-t}, \textbf{SAR-s} and \textbf{SAR}, donated as the token-shifted predictive entropy, sentence-shifted predictive entropy, and both token- and
sentence-shifted predictive entropy respectively.

\begin{table*}[tb]
\centering
\resizebox{\linewidth}{!}{
\begin{tabular}{lccc|ccc|ccc|ccc}
\toprule
 & \multicolumn{6}{c}{\textbf{TriviaQA}} & \multicolumn{6}{c}{\textbf{SciQA}} \\
\cmidrule (lr){2-7}\cmidrule (lr){8-13}
 & \multicolumn{3}{c}{\textbf{AUROC($\uparrow$)}} & \multicolumn{3}{c}{\textbf{ECE($\downarrow$)}} & \multicolumn{3}{c}{\textbf{AUROC($\uparrow$)}} & \multicolumn{3}{c}{\textbf{ECE($\downarrow$)}}\\
\cmidrule (lr){2-4}\cmidrule (lr){5-7}\cmidrule (lr){8-10}\cmidrule (lr){11-13}
    \textbf{Method} & \textbf{Vanilla} & \textbf{+Corrector} & \textbf{Improv} & \textbf{Vanilla} & \textbf{+Corrector} & \textbf{Improv}  & \textbf{Vanilla} & \textbf{+Corrector} & \textbf{Improv} & \textbf{Vanilla} & \textbf{+Corrector} & \textbf{Improv} \\
\midrule
\headercolor
\multicolumn{13}{c}{\textbf{OPT-6.7B}} \\
\textbf{LS} & 46.49  & \textbf{65.11}  & +18.62  & 72.71  & \textbf{41.76}  & -30.94  & 44.12  & \textbf{49.40}  & +5.29  & 76.38  & \textbf{32.78}  & -43.60 \\
\textbf{VC} & 60.41  & \textbf{70.55}  & +10.15  & 49.13  & \textbf{27.61}  & -21.52  & 51.69  & \textbf{56.55}  & +4.86  & 62.65  & \textbf{38.99}  & -23.66 \\
\textbf{P(True)} & 66.74  & \textbf{72.29}  & +5.84   & 45.00  & \textbf{32.63}  & -12.80  & 56.12  & \textbf{59.49}  & +3.37  & 58.79  & \textbf{34.52}  & -24.27 \\
\textbf{PE} & 56.36  & \textbf{66.62}  & +10.25  & 42.39  & \textbf{20.28}  & -22.12  & 50.07  & \textbf{56.02}  & +5.95  & 62.05  & \textbf{36.92}  & -25.13 \\
\textbf{LN-PE} & 78.37  & \textbf{79.93}  & +1.57   & 32.29  & \textbf{20.80}  & -11.49  & 60.88  & \textbf{64.23}  & +3.35  & 49.52  & \textbf{34.68}  & -14.84 \\
\textbf{SE} & 80.66  & \textbf{81.00}  & +0.34   & 36.64  & \textbf{27.05}  & -9.59   & 64.52  & \textbf{66.15}  & +1.63  & 52.66  & \textbf{42.23}  & -10.43 \\
\textbf{SAR-t} & 78.24  & \textbf{80.21}  & +1.97   & 40.14  & \textbf{37.85}  & -2.30   & 60.00  & \textbf{63.75}  & +3.74  & 45.33  & \textbf{44.19}  & -1.14 \\
\textbf{SAR-s} & 51.77  & \textbf{55.83}  & +4.06   & 53.78  & \textbf{49.65}  & -4.13   & 53.20  & \textbf{54.15}  & +0.95  & 76.21  & \textbf{34.83}  & -41.38 \\
\textbf{SAR} & 75.32  & \textbf{78.67}  & +3.35   & 40.61  & \textbf{31.02}  & -9.59   & 60.04  & \textbf{62.72}  & +2.68  & 49.40  & \textbf{38.99}  & -10.41 \\

\midrule
\headercolor
\multicolumn{13}{c}{\textbf{LLaMA-3-8B-Instruct}} \\
\textbf{LS} & 19.57  & \textbf{69.82}  & +50.25  & 70.25  & \textbf{7.41}  & -62.84  & 53.67  & \textbf{65.38}  & +11.71  & 38.64  & \textbf{18.19}  & -20.45 \\
\textbf{VC} & 62.34  & \textbf{74.89}  & +12.55  & 23.41  & \textbf{16.78}  & -6.63   & 68.22  & \textbf{72.15}  & +3.93   & 31.88  & \textbf{19.47}  & -12.36 \\
\textbf{P(True)} & 57.14  & \textbf{72.29}  & +15.15  & 24.67  & \textbf{19.84}  & -4.83   & 65.63  & \textbf{71.41}  & +5.78   & 34.56  & \textbf{31.92}  & -2.64  \\
\textbf{PE} & 64.52  & \textbf{69.76}  & +5.25   & 21.38  & \textbf{17.24}  & -4.13   & 66.54  & \textbf{67.98}  & +1.44   & 40.67  & \textbf{34.07}  & -6.60  \\
\textbf{LN-PE} & 72.55  & \textbf{74.79}  & +2.24  & 14.31  & \textbf{11.53}  & -2.79  & 69.48  & \textbf{71.56}  & +2.08  & 29.38  & \textbf{23.76}  & -5.62  \\
\textbf{SE} & 80.92  & \textbf{82.12}  & +1.20  & 13.07  & \textbf{12.76}  & -0.31  & 71.59  & \textbf{72.93 } & +1.34  & 30.54  & \textbf{25.23}  & -5.30  \\
\textbf{SAR-t} & 79.55  & \textbf{79.93}  & +0.38  & 16.40  & \textbf{13.70}  & -2.70  & 72.26  & \textbf{73.87}  & +1.61  & 30.37  & \textbf{26.81}  & -3.56  \\
\textbf{SAR-s} & 69.87  & \textbf{77.09}  & +2.95  & 23.17  & \textbf{20.00}  & -3.17  & 74.96  & \textbf{75.72}  & +0.76  & 38.54  & \textbf{36.18 } & -2.37  \\
\textbf{SAR} & 80.92  & \textbf{81.90}  & +0.98  & 16.17  & \textbf{13.76 } & -2.41  & 73.88  & \textbf{75.19}  & +1.31  & 28.97  & \textbf{25.60}  & -3.37  \\
\bottomrule
\end{tabular}
}
\caption{AUROC and ECE scores (\%) on the TriviaQA and SciQA datasets obtained by applying the \textit{Corrector} to existing uncertainty estimation methods. LS denotes the Lexical Similarity method. VC denotes the Verbal Confidence method. PE denote the Predictive Entropy method. LN-PE denotes the Length-normalized Predictive Entropy method. SE denote the Semantic Entropy. SAR-t refers to the token-level version of the SAR method, while SAR-s denotes the sentence-level version.}
\label{tab:main_res}
\end{table*}
\label{analysis}

\subsubsection{Implementation Details}
\paragraph{Dataset Splitting} For the TriviaQA dataset, we randomly selected 5,000 samples from the training set for data crafting and corrector training. For datasets with limited data, SciQA, we utilized the entire training set. We then used half of the test set to search for the optimal hyperparameter w, while the other half was employed to evaluate the method's effectiveness.

% \paragraph{Hyperparameter} For each dataset and model pair, we train a corresponding \textit{Corrector}, which is universally applicable across various methods. Additionally, for every method, dataset, and model combination, we derive the weight using the development set respectively. The sensitivity analysis of hyperparameter $w^∗$ and its configuration within the cross-domain experiments are elaborated upon in Appendix \ref{app:detail_of_w}.

\paragraph{Hyperparameter} For each dataset and model pair, we train a corresponding \textit{Corrector}, which is universally applicable across various methods. Additionally, for every method, dataset, and model combination, we derive the weight using the development set respectively. The sensitivity analysis of hyperparameter $w^*$ and its configuration within the cross-domain experiments are elaborated upon in Appendix \ref{app:detail_of_w}.
% \section{Result \& Analysis}

\begin{figure*}
    \centering
    \includegraphics[width=0.85\textwidth]{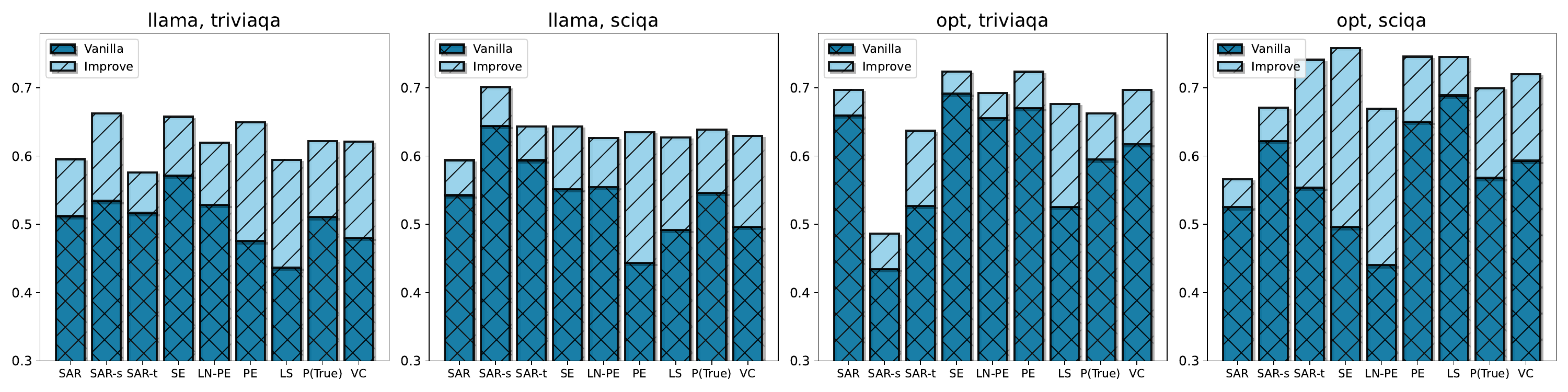}
    \caption{The performance gains of using the \textit{Corrector} to adjust the uncertainty scores for various methods on the datasets of TriviaQA and SciQA, and the target models of LLaMA-3-8B-Instruct and OPT-6.7B, are evaluated in terms of F1 score.}
    \label{fig:f1}
\end{figure*}

% In this section, we begin by evaluating various existing uncertainty estimation methods from both classification and calibration views .
% Following this, we conducted a straightforward cross-analysis of various methods and discovered that xxxx.
% Inspiringly, by applying our \textit{Corrector}, we observed a consistent and obvious improvement across multiple methods, models, and datasets.

% focusing on their limitations, particularly. We explored multiple key questions related to the impact of inherent flaws in large models, the complementarity of different uncertainty estimation methods, and the potential of external perspectives for improving model performance.

% \textbf{Do mainstream baselines based on internal model information suffer from inherent flaws in large models?}

\subsection{Main Result}
We has evaluated existing methods in Section \ref{sec:preliminart_study}
and found that there still remains a large performance gap between existing methods to achieve the harmonized uncertainty estimation. 
% These findings highlight considerable room for refinement in uncertainty estimation.
In this part, we present the performance of CUE\raisebox{-0.4ex}{\includegraphics[width=0.028\textwidth, keepaspectratio]{picture/light-bulb_1f4a1.png}} from both classification and calibration views, demonstrating that integrating a \textit{Corrector} with existing UE methods significantly enhances uncertainty estimation across multiple dimensions, including classification indication, precision-recall balance, and calibration.

\paragraph{Classification View}
\label{sec:comp}

As illustrated in Table \ref{tab:main_res}, the \textit{Corrector} has resulted in significant improvements, with an average AUROC score increase of 0.27 for TriviaQA and 0.09 for SciQA. Even when applied to challenging methods such as SE and SAR, the \textit{Corrector} boosts AUROC scores by 0.01 to 0.03. 
% Additionally, the peak AUROC scores achieved are 0.82 for TriviaQA and 0.75 for SciQA. 
% Since AUROC reflects UE method's ability to score instances for which the target model responds unreliably higher than those it responds to reliably, 
% these improvements indicate that the deployment of the \textit{Corrector} \textbf{enhances the overall indicative capacity of the uncertainty scores, thereby more effectively assisting users in determining whether to trust the model's responses.}
Since AUROC reflects the UE method’s ability to assign higher scores to instances where the target model responds unreliably compared to those it responds to reliably, these improvements indicate that the deployment of the \textit{Corrector} \textbf{enhances the overall indicative capacity of the uncertainty scores, making it more effective for users in determining whether to trust the model’s responses.}
% , thereby improving the reliability of uncertainty estimation.
% Even when combining with challenging baselines such as SAR, the \textit{Corrector} achieved enhancements of xx and xx.
% universally boosts performance in every methods for uncertainty estimation, across all datasets and models.

% for uncertainty scores derived from various methods
Furthermore, as illustrated in Figure \ref{fig:f1}, the F1 score is also boosted by the \textit{Corrector}, achieving an average increase of 38.97\%. This notable improvement demonstrates the \textit{Corrector}'s ability to help balance precision and recall, effectively mitigating the polarization tendency in the uncertainty scores observed in previous methods. 

\paragraph{Calibration View}
\label{sec:calibrate}

% As explained in Section \ref{metric}, while calibration typically focuses on confidence, we treat confidence and uncertainty as two sides of the same coin when calculating ECE.
% In a group of questions with an uncertainty score of 0.8, corresponding to a confidence level of 0.2, it can be inferred that there is a 20\% likelihood that this group of questions will be answered correctly by target model.
% In our discussion on calibration, we adhere to the experimental settings as previously outlined in Section \ref{sec:comp}.

% demonstrate how well the predictions align with the true outcomes across different probability thresholds

% , including same target models, datasets and baselines.

% As shown in Table \ref{tab:main_res},  it appears that prior methods have overlooked the calibration aspect,  resulting in relatively poor performance in terms of ECE scores.

\begin{figure*}[tb]
    \centering
    \includegraphics[width=1\textwidth]{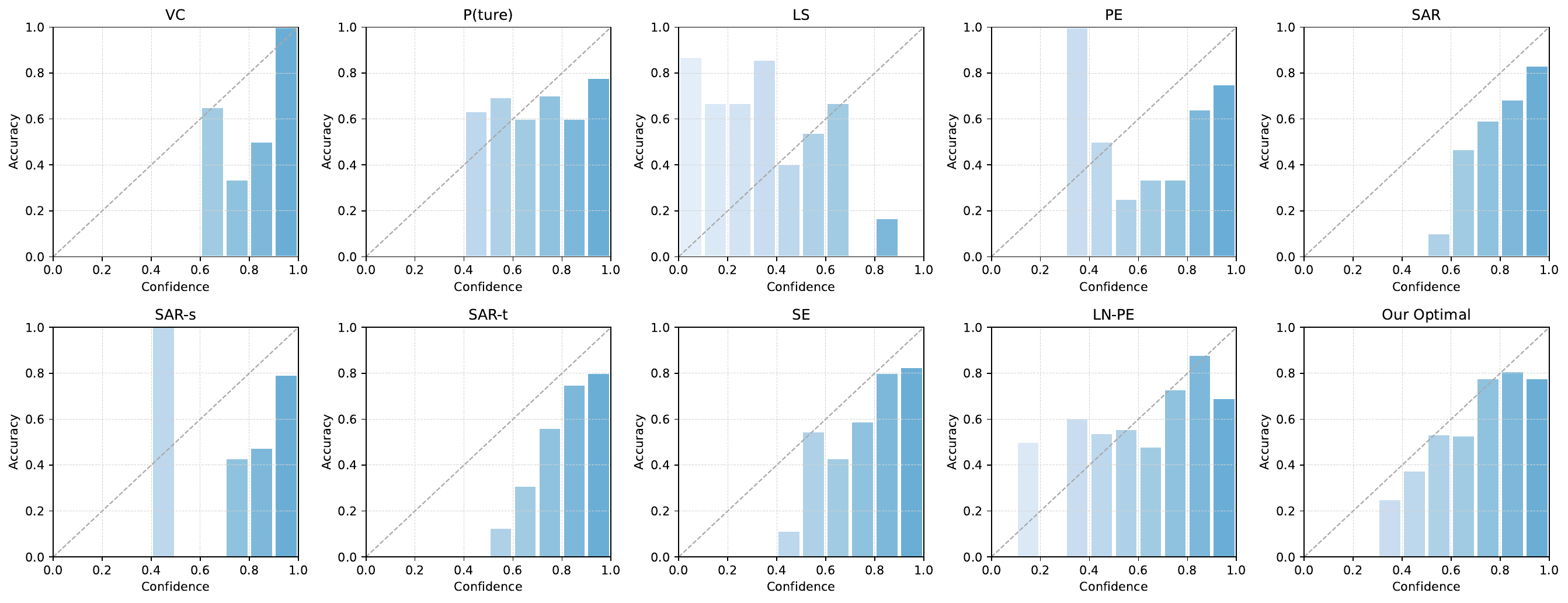}
    \caption{Calibration Plots. These plots depict the relationship between predicted confidence and observed frequencies. The diagonal line represents perfect calibration, where predicted confidence aligns precisely with actual outcomes. Bars extending above the diagonal indicate underestimation of confidence, while bars below the diagonal reflect overestimation. The final plot highlights the optimal calibration performance achieved through our \textit{Corrector}.}
    \label{fig:cal}
\end{figure*}

% which reflect inadequate absolute uncertainty
% These high ECE scores highlight a discrepancy between the uncertainty scores and actual conditions.
\textbf{Although calibration is not the direct training objective of our \textit{Corrector}, its application yields favorable calibration results.}
When employing the OPT-6.7B model as the target, we observed average ECE reductions of 0.34 on TriviaQA and 0.21 on SciQA. With the LLaMA-3-8B-Instruct model as the target, the reductions are 0.11 and 0.07, respectively—still considerable.
To further illustrate the calibration performance facilitated by the \textit{Corrector}, we provide calibration plots in Figure \ref{fig:cal}. 

We also conducted extensive experiments on the generalization performance of the \textit{Corrector}. Due to space constraints, we put the detailed results and analysis in the Appendix \ref{app:gene}.

In summary, integrating the \textit{Corrector} helps achieve harmonized uncertainty estimation. With the \textit{Corrector},  we can improve the reliability of uncertainty scores and alignment with the actual performance of the model.
The analysis of the pure Corrector's performance can be found in Appendix \ref{app:pure_corrector}.

% In summary, integrating the \textit{Corrector} with existing UE methods enables harmonized uncertainty estimation. This improves the reliability of uncertainty scores and their alignment with the model's actual performance.

 % including classification indication, precision-recall balance, and calibration
% These results show that the \textit{Corrector} 

% In summary, integrating the \textit{Corrector} significantly enhances uncertainty estimation across multiple dimensions and helps achieve harmonized xxx. With the Corrector,  improving the reliability of uncertainty scores and alignment with the actual performance of the model.

% , which clearly demonstrate the substantial enhancements in calibration accuracy facilitated by our \textit{Corrector}.

% \textbf{(4) Existing uncertainty estimation methods generally struggle with absolute uncertainty.}

% We perform a weighted sum of the uncertainty scores obtained from each method, determining the optimal weight on the development set and evaluating the combined model on the test set. We then compare the performance of the weighted combination with that of the best-performing individual methods from the pair

\subsection{Ablation Study}
% \vspace{-12pt}
% We infer that the insufficient complementarity among current methods may be attributed to two main reasons. 
% For methods within the same category, derived from or are improvements upon previous ones, enhancing performance by optimizing existing mechanisms. In such cases, the improved methods essentially integrate the strengths of their predecessors while addressing or circumventing their weaknesses. 

% many methods are derived from or are improvements upon previous ones, enhancing performance by introducing additional features or optimizing existing mechanisms. 
% In such cases, the improved methods essentially integrate the strengths of their predecessors while addressing or circumventing their weaknesses. 
% Consequently, these improved methods exhibit superior performance and dominate the overall effectiveness of uncertainty estimation.
% Furthermore, we observe that methods across different categories also lack complementarity. This may be due to the inherent instability in the performance of these methods and the significant logical overlap among them. This underscores the importance and necessity of incorporating external heterogeneous information introduced by \textit{Corrector}.

We conducted ablation studies to scrutinize the impact of the base model, the correction score formats and its acquisition methods.

\paragraph{Formats} We compared the efficacy of probabilistic values versus label values for correction.
As shown in Table \ref{tab:ablation}, probabilistic correction scores demonstrate clear superiority, as they allow finer-grained adjustments by leveraging a broader spectrum for integration.
Conversely, discrete values, such as 0 and 1, tend to introduce significant biases in the corrected uncertainty scores.

\begin{figure}[ht]
    \centering
    \includegraphics[width=0.5\textwidth]{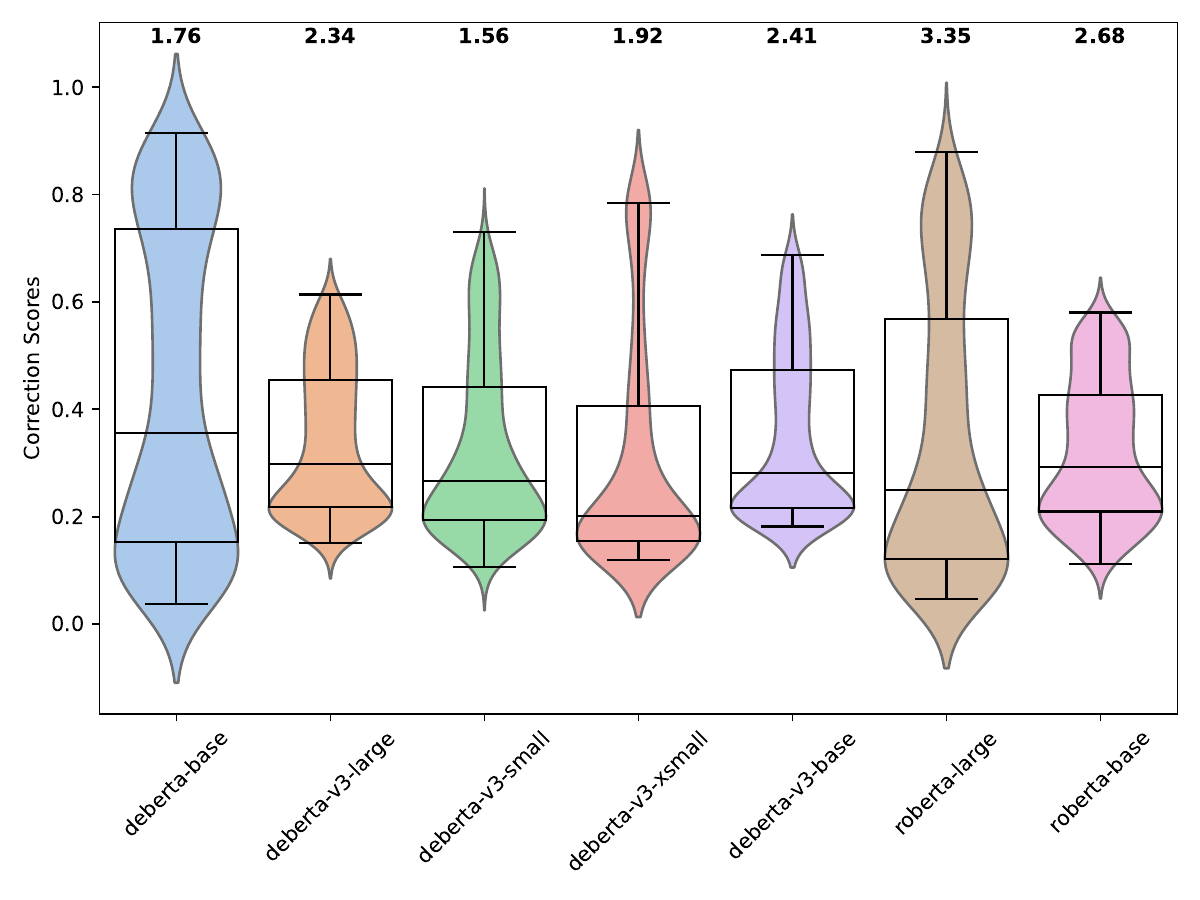}
    \caption{The overall AUROC score gains achieved by \textit{Correctors} trained on different base models across various UE methods on the SciQA dataset and Llama-3-8B-Instruct target model.
}
    \label{fig:base_model}
\end{figure}

\paragraph{Base Model} We utilized various encoder models as base models to train the \textit{Corrector} and assess the impact on correction performance. Specifically, we employed models from the RoBERTa series, including RoBERTa-base\footnote{\url{huggingface.co/FacebookAI/roberta-base}} and RoBERTa-large\footnote{\url{huggingface.co/FacebookAI/roberta-large}}, as well as models from the DeBERTa series, including DeBERTa-base\footnote{\url{huggingface.co/microsoft/deberta-base}}, DeBERTa-v3-large\footnote{\url{huggingface.co/microsoft/deberta-v3-large}}, DeBERTa-v3-base\footnote{\url{huggingface.co/microsoft/deberta-v3-base}}, DeBERTa-v3-small\footnote{\url{huggingface.co/microsoft/deberta-v3-small}}, and DeBERTa-v3-xsmall\footnote{\url{huggingface.co/microsoft/deberta-v3-xsmall}}. These models represent different types, series, and sizes. As illustrated in Figure \ref{fig:base_model}, more advanced, later-generation, and larger models yield superior results.

% \vspace{-20pt}

% \paragraph{Acquisition} We compare the correction scores obtained from a lightweight classifier with those estimated using GPT-4o. It’s important to note that our goal is not to rigorously assess the target model’s answers, but rather to predict its reliability in answering questions. Despite GPT-4o's strong performance in question answering, our results reveal that it is less effective than the classifier in directly predicting the reliability of target models when faced with questions.
% Additionally, as detailed in Section \ref{sec:inter-methed}, integrating uncertainty scores from different UE methods does not improve performance and may even degrade it. This underscores the \textit{Corrector}'s unique and powerful role as a complement, distinct from existing uncertainty estimation methods.

\paragraph{Acquisition} We compare correction scores from a lightweight classifier with those estimated using GPT-4o. We attempt not to rigorously assess the target model’s answers but to predict its reliability. Despite GPT-4's strong performance in question answering, our results show it is less effective than the classifier in directly predicting reliability of target models when faced with questions. Additionally, as detailed in Section \ref{sec:inter-methed}, combining uncertainty scores from different UE methods does not improve and may even degrade performance. This highlights the \textit{Corrector}'s unique role as a complement to existing UE methods.

We also conducted experiments to compare our \textit{Corrector} with other supervised methods specifically designed for uncertainty estimation. The experimental results demonstrated that our supervised pipeline offered significant advantages in enhancing uncertainty estimation. Detailed results and analysis can be found in Appendix \ref{app:comparative}.

\begin{table}[H]
% \small
    \centering
    \resizebox{\linewidth}{!}{
    \begin{tabular}{l|c|c}
    \toprule
    \textbf{Methods}&\textbf{AUROC ($\uparrow$)}&\textbf{ECE ($\downarrow$)}\\
    \midrule
    \textbf{Corrector} & 69.87  & 6.73 \\
    \textbf{Original Best}  & 80.92 & 11.53 \\
    \textbf{+Corrector Probability} &\textbf{ 82.12} & \textbf{10.46} \\
    \textbf{+Corrector Label}  & 80.92 & 11.53\\
    \textbf{+GPT-4o Score} &  80.92 & 11.53\\
    \bottomrule
    \end{tabular}}
    \caption{Ablation Study. LLaMA-3-8B-Instruct as the target model and TriviaQA as the test dataset. \textbf{Original Best} refers to the peak performance achieved by various baseline when the \textit{Corrector} is not incorporated.}
    \label{tab:ablation}
\end{table}

\section{Conclusion}

Our study highlights the limitations of current uncertainty estimation methods in terms of classification accuracy, precision-recall balance, and calibration. 
We introduce an innovative uncertainty score correction framework that utilizes a classifier as a \textit{Corrector} to refine these scores, ensuring alignment with the model's true task performance. This \textit{Corrector} integrates seamlessly with existing methods, enhancing their effectiveness. 
Extensive experiments validate that the \textit{Corrector} consistently improves performance across various metrics, data domains, and target models. Furthermore, our ablation study underscores the \textit{Corrector}'s capacity to provide substantial and heterogeneous improvements to existing uncertainty estimation techniques.

% We find that existing uncertainty estimation methods are often limited by the over-confidence and under-confidence inherent in LLMs, leading to inaccuracies in uncertainty estimation.
% To address these issues,  we propose an external insight-driven correction approach which enables seamless integration with existing uncertainty estimation methods.
% Our method offers a fresh perspective by leveraging external information to improve uncertainty estimation, 
% complementing traditional approaches that rely on the internal logic or outputs of the models.
% Through a bidirectional calibration method, we effectively enhance the accuracy of uncertainty estimates. 
% Our method has demonstrated superior performance across various models, methods, and datasets.
% We demonstrate that our method consistently outperforms existing approaches included in the Representative Baselines Set (RBS) and the Challenging Baselines Set (CBS), exhibiting significant improvements in both relative and absolute terms. Furthermore, we present comprehensive experimental evidence underscoring the robustness and generalizability of our approach across diverse data domains and target models.

\section*{Limitations}

Although the CUE method proposed in this paper demonstrates good performance, its dependence on labeled data and its generalization ability across different data domains and target models may be limitations.
%Although the CUE method proposed in the paper demonstrates good performance, its generalization ability across different data domains and target models may be limited. 
% When constructing the dataset for training the Corrector, it is necessary to accurately evaluate the responses generated by the target model. 
We only compared our method with works that have open-source code, which are often designed for white-box models. Therefore, the effectiveness of our method on black-box models has not been demonstrated through experiments. However, our method does not necessitate access to the inner states of target models, making it a general enhancement strategy for both black-box and white-box uncertainty estimation.

\section*{Ethics Statement}
In this study, we introduce a method for improving uncertainty estimation in the context of LLMs, which presents no immediate ethical concerns, but certain considerations must be addressed. 
Uncertainty estimation has significant potential to evaluate the reliability and safety of LLM outputs. 
However, this potential benefit comes with the risk that systematic mistakes in the uncertainty assessment could foster unfounded and misplaced confidence. Consequently, even re-calibrated uncertainty estimates should be interpreted cautiously, particularly in critical decision-making scenarios where the consequences of inaccuracies can be profound.

The datasets used in our experiment are publicly released and labeled through interaction with humans in English. In this process, user privacy is protected, and no personal information is contained in the dataset. The scientific artifacts that we used are available for research with permissive licenses. And the use of these artifacts in this paper is consistent with their intended use. Therefore, we believe that our research work meets the ethics of ACL.

\section*{Acknowledgements}
% Entries for the entire Anthology, followed by custom entries
This paper is supported by NSFC project 62476009.

\bibliography{custom}

\begin{thebibliography}{51}
\providecommand{\natexlab}[1]{#1}

\bibitem[{Allikivi et~al.(2024)Allikivi, J{\""a}rve, and Kull}]{allikivi2024cautious}
Mari-Liis Allikivi, Joonas J{\""a}rve, and Meelis Kull. 2024.
\newblock Cautious calibration in binary classification.
\newblock \emph{arXiv preprint arXiv:2408.05120}.

\bibitem[{Auer et~al.(2023)Auer, Barone, Bartz, Cortes, Jaradeh, Karras, Koubarakis, Mouromtsev, Pliukhin, Radyush, Shilin, Stocker, and Tsalapati}]{SciQA2023}
S{\"o}ren Auer, Dante A.~C. Barone, Cassiano Bartz, Eduardo~G. Cortes, Mohamad~Yaser Jaradeh, Oliver Karras, Manolis Koubarakis, Dmitry Mouromtsev, Dmitrii Pliukhin, Daniil Radyush, Ivan Shilin, Markus Stocker, and Eleni Tsalapati. 2023.
\newblock \href {https://doi.org/10.1038/s41598-023-33607-z} {The sciqa scientific question answering benchmark for scholarly knowledge}.
\newblock \emph{Scientific Reports}, 13(1):7240.

\bibitem[{Azaria and Mitchell(2023)}]{azaria2023internal}
Amos Azaria and Tom Mitchell. 2023.
\newblock The internal state of an llm knows when it's lying.
\newblock \emph{arXiv preprint arXiv:2304.13734}.

\bibitem[{Bakman et~al.(2024)Bakman, Yaldiz, Buyukates, Tao, Dimitriadis, and Avestimehr}]{bakman2024mars}
Yavuz~Faruk Bakman, Duygu~Nur Yaldiz, Baturalp Buyukates, Chenyang Tao, Dimitrios Dimitriadis, and Salman Avestimehr. 2024.
\newblock Mars: Meaning-aware response scoring for uncertainty estimation in generative llms.
\newblock \emph{arXiv preprint arXiv:2402.11756}.

\bibitem[{Becker and Soatto(2024)}]{becker2024cycles}
Evan Becker and Stefano Soatto. 2024.
\newblock Cycles of thought: Measuring llm confidence through stable explanations.
\newblock \emph{arXiv preprint arXiv:2406.03441}.

\bibitem[{Chaudhry et~al.(2024)Chaudhry, Thiagarajan, and Gorur}]{chaudhry2024finetuninglanguagemodelsemit}
Arslan Chaudhry, Sridhar Thiagarajan, and Dilan Gorur. 2024.
\newblock \href {https://arxiv.org/abs/2409.12180} {Finetuning language models to emit linguistic expressions of uncertainty}.
\newblock \emph{Preprint}, arXiv:2409.12180.

\bibitem[{Duan et~al.(2023)Duan, Cheng, Wang, Wang, Zavalny, Xu, Kailkhura, and Xu}]{duan2023shifting}
Jinhao Duan, Hao Cheng, Shiqi Wang, Chenan Wang, Alex Zavalny, Renjing Xu, Bhavya Kailkhura, and Kaidi Xu. 2023.
\newblock Shifting attention to relevance: Towards the uncertainty estimation of large language models.
\newblock \emph{arXiv preprint arXiv:2307.01379}.

\bibitem[{Dubey et~al.(2024)Dubey, Jauhri, Pandey, Kadian, Al-Dahle, Letman, Mathur, Schelten, Yang, Fan et~al.}]{dubey2024llama}
Abhimanyu Dubey, Abhinav Jauhri, Abhinav Pandey, Abhishek Kadian, Ahmad Al-Dahle, Aiesha Letman, Akhil Mathur, Alan Schelten, Amy Yang, Angela Fan, et~al. 2024.
\newblock The llama 3 herd of models.
\newblock \emph{arXiv preprint arXiv:2407.21783}.

\bibitem[{Fomicheva et~al.(2020)Fomicheva, Sun, Yankovskaya, Blain, Guzm{\'a}n, Fishel, Aletras, Chaudhary, and Specia}]{fomicheva2020unsupervised}
Marina Fomicheva, Shuo Sun, Lisa Yankovskaya, Fr{\'e}d{\'e}ric Blain, Francisco Guzm{\'a}n, Mark Fishel, Nikolaos Aletras, Vishrav Chaudhary, and Lucia Specia. 2020.
\newblock Unsupervised quality estimation for neural machine translation.
\newblock \emph{Transactions of the Association for Computational Linguistics}, 8:539--555.

\bibitem[{Gregor et~al.(2014)Gregor, Danihelka, Mnih, Blundell, and Wierstra}]{gregor2014deep}
Karol Gregor, Ivo Danihelka, Andriy Mnih, Charles Blundell, and Daan Wierstra. 2014.
\newblock Deep autoregressive networks.
\newblock In \emph{International Conference on Machine Learning}, pages 1242--1250. PMLR.

\bibitem[{Groot and Valdenegro-Toro(2024)}]{groot2024overconfidence}
Tobias Groot and Matias Valdenegro-Toro. 2024.
\newblock Overconfidence is key: Verbalized uncertainty evaluation in large language and vision-language models.
\newblock \emph{arXiv preprint arXiv:2405.02917}.

\bibitem[{Han et~al.(2024)Han, Li, Chen, Shi, Du, Xiao, Liang, and Lin}]{han2024enhancing}
Haixia Han, Tingyun Li, Shisong Chen, Jie Shi, Chengyu Du, Yanghua Xiao, Jiaqing Liang, and Xin Lin. 2024.
\newblock Enhancing confidence expression in large language models through learning from past experience.
\newblock \emph{arXiv preprint arXiv:2404.10315}.

\bibitem[{He et~al.(2021{\natexlab{a}})He, Gao, and Chen}]{he2021debertav3}
Pengcheng He, Jianfeng Gao, and Weizhu Chen. 2021{\natexlab{a}}.
\newblock \href {https://arxiv.org/abs/2111.09543} {Debertav3: Improving deberta using electra-style pre-training with gradient-disentangled embedding sharing}.
\newblock \emph{Preprint}, arXiv:2111.09543.

\bibitem[{He et~al.(2021{\natexlab{b}})He, Liu, Gao, and Chen}]{he2021deberta}
Pengcheng He, Xiaodong Liu, Jianfeng Gao, and Weizhu Chen. 2021{\natexlab{b}}.
\newblock \href {https://openreview.net/forum?id=XPZIaotutsD} {Deberta: Decoding-enhanced bert with disentangled attention}.
\newblock In \emph{International Conference on Learning Representations}.

\bibitem[{Ji et~al.(2024)Ji, Chen, Ishii, Cahyawijaya, Bang, Wilie, and Fung}]{ji2024llm}
Ziwei Ji, Delong Chen, Etsuko Ishii, Samuel Cahyawijaya, Yejin Bang, Bryan Wilie, and Pascale Fung. 2024.
\newblock Llm internal states reveal hallucination risk faced with a query.
\newblock \emph{arXiv preprint arXiv:2407.03282}.

\bibitem[{Jin et~al.(2020)Jin, Pan, Oufattole, Weng, Fang, and Szolovits}]{jin2020diseasedoespatienthave}
Di~Jin, Eileen Pan, Nassim Oufattole, Wei-Hung Weng, Hanyi Fang, and Peter Szolovits. 2020.
\newblock \href {https://arxiv.org/abs/2009.13081} {What disease does this patient have? a large-scale open domain question answering dataset from medical exams}.
\newblock \emph{Preprint}, arXiv:2009.13081.

\bibitem[{Joshi et~al.(2017)Joshi, Choi, Weld, and Zettlemoyer}]{joshi2017triviaqa}
Mandar Joshi, Eunsol Choi, Daniel~S Weld, and Luke Zettlemoyer. 2017.
\newblock Triviaqa: A large scale distantly supervised challenge dataset for reading comprehension.
\newblock \emph{arXiv preprint arXiv:1705.03551}.

\bibitem[{Kadavath et~al.(2022)Kadavath, Conerly, Askell, Henighan, Drain, Perez, Schiefer, Hatfield-Dodds, DasSarma, Tran-Johnson et~al.}]{kadavath2022language}
Saurav Kadavath, Tom Conerly, Amanda Askell, Tom Henighan, Dawn Drain, Ethan Perez, Nicholas Schiefer, Zac Hatfield-Dodds, Nova DasSarma, Eli Tran-Johnson, et~al. 2022.
\newblock Language models (mostly) know what they know.
\newblock \emph{arXiv preprint arXiv:2207.05221}.

\bibitem[{Kamath et~al.(2020)Kamath, Jia, and Liang}]{kamath2020selective}
Amita Kamath, Robin Jia, and Percy Liang. 2020.
\newblock Selective question answering under domain shift.
\newblock \emph{arXiv preprint arXiv:2006.09462}.

\bibitem[{Kapoor et~al.(2024)Kapoor, Gruver, Roberts, Pal, Dooley, Goldblum, and Wilson}]{kapoor-etal-2024-calibration}
Sanyam Kapoor, Nate Gruver, Manley Roberts, Arka Pal, Samuel Dooley, Micah Goldblum, and Andrew Wilson. 2024.
\newblock \href {https://aclanthology.org/2024.uncertainlp-1.1/} {Calibration-tuning: Teaching large language models to know what they don`t know}.
\newblock In \emph{Proceedings of the 1st Workshop on Uncertainty-Aware NLP (UncertaiNLP 2024)}, pages 1--14, St Julians, Malta. Association for Computational Linguistics.

\bibitem[{Kossen et~al.(2024)Kossen, Han, Razzak, Schut, Malik, and Gal}]{kossen2024semantic}
Jannik Kossen, Jiatong Han, Muhammed Razzak, Lisa Schut, Shreshth Malik, and Yarin Gal. 2024.
\newblock Semantic entropy probes: Robust and cheap hallucination detection in llms.
\newblock \emph{arXiv preprint arXiv:2406.15927}.

\bibitem[{Kuhn et~al.(2023)Kuhn, Gal, and Farquhar}]{kuhn2023semantic}
Lorenz Kuhn, Yarin Gal, and Sebastian Farquhar. 2023.
\newblock Semantic uncertainty: Linguistic invariances for uncertainty estimation in natural language generation.
\newblock \emph{arXiv preprint arXiv:2302.09664}.

\bibitem[{Li et~al.(2024{\natexlab{a}})Li, Patel, Vi{\'e}gas, Pfister, and Wattenberg}]{li2024inference}
Kenneth Li, Oam Patel, Fernanda Vi{\'e}gas, Hanspeter Pfister, and Martin Wattenberg. 2024{\natexlab{a}}.
\newblock Inference-time intervention: Eliciting truthful answers from a language model.
\newblock \emph{Advances in Neural Information Processing Systems}, 36.

\bibitem[{Li et~al.(2024{\natexlab{b}})Li, Wang, Feng, Zhu, Wang, and Chua}]{li2024think}
Moxin Li, Wenjie Wang, Fuli Feng, Fengbin Zhu, Qifan Wang, and Tat-Seng Chua. 2024{\natexlab{b}}.
\newblock Think twice before assure: Confidence estimation for large language models through reflection on multiple answers.
\newblock \emph{arXiv preprint arXiv:2403.09972}.

\bibitem[{Lin et~al.(2023)Lin, Trivedi, and Sun}]{lin2023generating}
Zhen Lin, Shubhendu Trivedi, and Jimeng Sun. 2023.
\newblock Generating with confidence: Uncertainty quantification for black-box large language models.
\newblock \emph{arXiv preprint arXiv:2305.19187}.

\bibitem[{Liu et~al.(2024)Liu, Pan, Li, and Chen}]{liu2024uncertainty}
Linyu Liu, Yu~Pan, Xiaocheng Li, and Guanting Chen. 2024.
\newblock Uncertainty estimation and quantification for llms: A simple supervised approach.
\newblock \emph{arXiv preprint arXiv:2404.15993}.

\bibitem[{Liu(2019)}]{liu2019roberta}
Yinhan Liu. 2019.
\newblock Roberta: A robustly optimized bert pretraining approach.
\newblock \emph{arXiv preprint arXiv:1907.11692}.

\bibitem[{Loquercio et~al.(2020)Loquercio, Segu, and Scaramuzza}]{loquercio2020general}
Antonio Loquercio, Mattia Segu, and Davide Scaramuzza. 2020.
\newblock A general framework for uncertainty estimation in deep learning.
\newblock \emph{IEEE Robotics and Automation Letters}, 5(2):3153--3160.

\bibitem[{Madhusudhan et~al.(2024)Madhusudhan, Madhusudhan, Yadav, and Hashemi}]{madhusudhan2024llms}
Nishanth Madhusudhan, Sathwik~Tejaswi Madhusudhan, Vikas Yadav, and Masoud Hashemi. 2024.
\newblock Do llms know when to not answer? investigating abstention abilities of large language models.
\newblock \emph{arXiv preprint arXiv:2407.16221}.

\bibitem[{Malinin and Gales(2020)}]{malinin2020uncertainty}
Andrey Malinin and Mark Gales. 2020.
\newblock Uncertainty estimation in autoregressive structured prediction.
\newblock \emph{arXiv preprint arXiv:2002.07650}.

\bibitem[{Ni et~al.(2024)Ni, Bi, Yu, and Guo}]{ni2024large}
Shiyu Ni, Keping Bi, Lulu Yu, and Jiafeng Guo. 2024.
\newblock Are large language models more honest in their probabilistic or verbalized confidence?
\newblock \emph{arXiv preprint arXiv:2408.09773}.

\bibitem[{Nikitin et~al.(2024)Nikitin, Kossen, Gal, and Marttinen}]{nikitin2024kernel}
Alexander Nikitin, Jannik Kossen, Yarin Gal, and Pekka Marttinen. 2024.
\newblock Kernel language entropy: Fine-grained uncertainty quantification for llms from semantic similarities.
\newblock \emph{arXiv preprint arXiv:2405.20003}.

\bibitem[{OpenAI(2023)}]{openai:2023gpt4}
OpenAI. 2023.
\newblock \href {https://doi.org/10.48550/ARXIV.2303.08774} {{GPT-4} technical report}.
\newblock \emph{CoRR}, abs/2303.08774.

\bibitem[{Ouyang et~al.(2022)Ouyang, Wu, Jiang, Almeida, Wainwright, Mishkin, Zhang, Agarwal, Slama, Ray, Schulman, Hilton, Kelton, Miller, Simens, Askell, Welinder, Christiano, Leike, and Lowe}]{Ouyang0JAWMZASR22}
Long Ouyang, Jeffrey Wu, Xu~Jiang, Diogo Almeida, Carroll~L. Wainwright, Pamela Mishkin, Chong Zhang, Sandhini Agarwal, Katarina Slama, Alex Ray, John Schulman, Jacob Hilton, Fraser Kelton, Luke Miller, Maddie Simens, Amanda Askell, Peter Welinder, Paul~F. Christiano, Jan Leike, and Ryan Lowe. 2022.
\newblock \href {http://papers.nips.cc/paper\_files/paper/2022/hash/b1efde53be364a73914f58805a001731-Abstract-Conference.html} {Training language models to follow instructions with human feedback}.
\newblock In \emph{Advances in Neural Information Processing Systems 35: Annual Conference on Neural Information Processing Systems 2022, NeurIPS 2022, New Orleans, LA, USA, November 28 - December 9, 2022}.

\bibitem[{Pal et~al.(2022)Pal, Umapathi, and Sankarasubbu}]{pal2022medmcqa}
Ankit Pal, Logesh~Kumar Umapathi, and Malaikannan Sankarasubbu. 2022.
\newblock Medmcqa: A large-scale multi-subject multi-choice dataset for medical domain question answering.
\newblock In \emph{Conference on health, inference, and learning}, pages 248--260. PMLR.

\bibitem[{Papadopoulos and Yeung(2001)}]{papadopoulos2001uncertainty}
Christos~E Papadopoulos and Hoi Yeung. 2001.
\newblock Uncertainty estimation and monte carlo simulation method.
\newblock \emph{Flow Measurement and Instrumentation}, 12(4):291--298.

\bibitem[{Pedapati et~al.(2024)Pedapati, Dhurandhar, Ghosh, Dan, and Sattigeri}]{pedapati2024large}
Tejaswini Pedapati, Amit Dhurandhar, Soumya Ghosh, Soham Dan, and Prasanna Sattigeri. 2024.
\newblock Large language model confidence estimation via black-box access.
\newblock \emph{arXiv preprint arXiv:2406.04370}.

\bibitem[{Shen et~al.(2024)Shen, Das, Greenewald, Sattigeri, Wornell, and Ghosh}]{shen2024thermometeruniversalcalibrationlarge}
Maohao Shen, Subhro Das, Kristjan Greenewald, Prasanna Sattigeri, Gregory Wornell, and Soumya Ghosh. 2024.
\newblock \href {https://arxiv.org/abs/2403.08819} {Thermometer: Towards universal calibration for large language models}.
\newblock \emph{Preprint}, arXiv:2403.08819.

\bibitem[{Tao et~al.(2023)Tao, Dong, and Xu}]{tao2023dual}
Linwei Tao, Minjing Dong, and Chang Xu. 2023.
\newblock Dual focal loss for calibration.
\newblock In \emph{International Conference on Machine Learning}, pages 33833--33849. PMLR.

\bibitem[{Tao et~al.(2024)Tao, Yao, Ding, Xie, Cao, Sun, Gao, Shen, and Ding}]{tao2024trust}
Shuchang Tao, Liuyi Yao, Hanxing Ding, Yuexiang Xie, Qi~Cao, Fei Sun, Jinyang Gao, Huawei Shen, and Bolin Ding. 2024.
\newblock When to trust llms: Aligning confidence with response quality.
\newblock \emph{arXiv preprint arXiv:2404.17287}.

\bibitem[{Touvron et~al.(2023{\natexlab{a}})Touvron, Lavril, Izacard, Martinet, Lachaux, Lacroix, Rozi{\`{e}}re, Goyal, Hambro, Azhar, Rodriguez, Joulin, Grave, and Lample}]{Hugo:2023llama}
Hugo Touvron, Thibaut Lavril, Gautier Izacard, Xavier Martinet, Marie{-}Anne Lachaux, Timoth{\'{e}}e Lacroix, Baptiste Rozi{\`{e}}re, Naman Goyal, Eric Hambro, Faisal Azhar, Aur{\'{e}}lien Rodriguez, Armand Joulin, Edouard Grave, and Guillaume Lample. 2023{\natexlab{a}}.
\newblock \href {https://doi.org/10.48550/arXiv.2302.13971} {Llama: Open and efficient foundation language models}.
\newblock \emph{CoRR}, abs/2302.13971.

\bibitem[{Touvron et~al.(2023{\natexlab{b}})Touvron, Martin, Stone, Albert, Almahairi, Babaei et~al.}]{Hugo:2023llama2}
Hugo Touvron, Louis Martin, Kevin Stone, Peter Albert, Amjad Almahairi, Yasmine Babaei, et~al. 2023{\natexlab{b}}.
\newblock \href {https://doi.org/10.48550/ARXIV.2307.09288} {Llama 2: Open foundation and fine-tuned chat models}.
\newblock \emph{CoRR}, abs/2307.09288.

\bibitem[{Ulmer et~al.(2024)Ulmer, Gubri, Lee, Yun, and Oh}]{ulmer2024calibratinglargelanguagemodels}
Dennis Ulmer, Martin Gubri, Hwaran Lee, Sangdoo Yun, and Seong~Joon Oh. 2024.
\newblock \href {https://arxiv.org/abs/2403.05973} {Calibrating large language models using their generations only}.
\newblock \emph{Preprint}, arXiv:2403.05973.

\bibitem[{Vazhentsev et~al.(2023)Vazhentsev, Kuzmin, Tsvigun, Panchenko, Panov, Burtsev, and Shelmanov}]{vazhentsev2023hybrid}
Artem Vazhentsev, Gleb Kuzmin, Akim Tsvigun, Alexander Panchenko, Maxim Panov, Mikhail Burtsev, and Artem Shelmanov. 2023.
\newblock Hybrid uncertainty quantification for selective text classification in ambiguous tasks.
\newblock In \emph{Proceedings of the 61st Annual Meeting of the Association for Computational Linguistics (Volume 1: Long Papers)}, pages 11659--11681.

\bibitem[{Wachter et~al.(2024)Wachter, Mittelstadt, and Russell}]{wachter2024large}
Sandra Wachter, Brent Mittelstadt, and Chris Russell. 2024.
\newblock Do large language models have a legal duty to tell the truth?
\newblock \emph{Royal Society Open Science}, 11(8):240197.

\bibitem[{Wang et~al.(2024)Wang, Duan, Yuan, Chen, Chen, Yao, Zhang, Wang, Xu, and Shi}]{wang2024word}
Zhiyuan Wang, Jinhao Duan, Chenxi Yuan, Qingyu Chen, Tianlong Chen, Huaxiu Yao, Yue Zhang, Ren Wang, Kaidi Xu, and Xiaoshuang Shi. 2024.
\newblock Word-sequence entropy: Towards uncertainty estimation in free-form medical question answering applications and beyond.
\newblock \emph{arXiv preprint arXiv:2402.14259}.

\bibitem[{Xiong et~al.(2023)Xiong, Hu, Lu, Li, Fu, He, and Hooi}]{xiong2023can}
Miao Xiong, Zhiyuan Hu, Xinyang Lu, Yifei Li, Jie Fu, Junxian He, and Bryan Hooi. 2023.
\newblock Can llms express their uncertainty? an empirical evaluation of confidence elicitation in llms.
\newblock \emph{arXiv preprint arXiv:2306.13063}.

\bibitem[{Yaldiz et~al.(2024)Yaldiz, Bakman, Buyukates, Tao, Ramakrishna, Dimitriadis, and Avestimehr}]{yaldiz2024not}
Duygu~Nur Yaldiz, Yavuz~Faruk Bakman, Baturalp Buyukates, Chenyang Tao, Anil Ramakrishna, Dimitrios Dimitriadis, and Salman Avestimehr. 2024.
\newblock Do not design, learn: A trainable scoring function for uncertainty estimation in generative llms.
\newblock \emph{arXiv preprint arXiv:2406.11278}.

\bibitem[{Yang et~al.(2024)Yang, Chen, and Pitas}]{yang2024just}
Adam Yang, Chen Chen, and Konstantinos Pitas. 2024.
\newblock Just rephrase it! uncertainty estimation in closed-source language models via multiple rephrased queries.
\newblock \emph{arXiv preprint arXiv:2405.13907}.

\bibitem[{Zhang et~al.(2022)Zhang, Roller, Goyal, Artetxe, Chen, Chen, Dewan, Diab, Li, Lin et~al.}]{zhang2022opt}
Susan Zhang, Stephen Roller, Naman Goyal, Mikel Artetxe, Moya Chen, Shuohui Chen, Christopher Dewan, Mona Diab, Xian Li, Xi~Victoria Lin, et~al. 2022.
\newblock Opt: Open pre-trained transformer language models.
\newblock \emph{arXiv preprint arXiv:2205.01068}.

\bibitem[{Zhang et~al.(2023)Zhang, Li, Cui, Cai, Liu, Fu, Huang, Zhao, Zhang, Chen et~al.}]{zhang2023siren}
Yue Zhang, Yafu Li, Leyang Cui, Deng Cai, Lemao Liu, Tingchen Fu, Xinting Huang, Enbo Zhao, Yu~Zhang, Yulong Chen, et~al. 2023.
\newblock Siren's song in the ai ocean: a survey on hallucination in large language models.
\newblock \emph{arXiv preprint arXiv:2309.01219}.

\end{thebibliography}

\appendix

\section{Appendix}
\label{sec:appendix}
\subsection{Related Work}
\label{app:more_work}

% Prior research on uncertainty estimation has primarily concentrated on scenarios such as classification or regression using standard uncertainty techniques, which are not directly applicable to LLMs. Therefore, there has been considerable interest in developing uncertainty estimation methods specifically for LLMs. These methods can be broadly categorized into logit-based methods, verbal confidence methods, and internal state-based methods and consistency-based method.

Uncertainty estimation methods for LLMs have gained significant attention, with approaches  can be broadly categorized into logit-based methods, verbalized methods, consistency-based methods, and internal state-based methods..

\textbf{Logit-based methods}
Logit-based methods are the most widely used and effective approaches in uncertainty estimation. 
% They derive uncertainty scores by converting model output logits using entropy-based techniques.
As a foundational method, Predictive Entropy (PE)~\citep{malinin2020uncertainty}, defines total uncertainty as the entropy of the output logits distribution.
% and is widely adopted in subsequent methods
% Predictive Entropy (PE)~\citep{malinin2020uncertainty}, the foundational logit-based uncertainty estimation method, defines total uncertainty as the entropy of the output distribution, which has been widely adopted by subsequent methods.
%Uncertainty estimation in autoregressive structured prediction
After that, researchers proposed a series of methods based on the inherent characteristics of natural language generation to improve upon PE methods.
\citet{kuhn2023semantic} introduced semantic entropy (SE) that estimates uncertainty by marginalizing over semantically-equivalent samples in NLG tasks. 
%Semantic uncertainty: Linguistic invariances for uncertainty estimation in natural language generation
In the similar framework, \citet{nikitin2024kernel} employed positive semi-definite kernels and von Neumann entropy to capture semantic similarities. 
%Kernel Language Entropy: Fine-grained Uncertainty Quantification for LLMs from Semantic Similarities
Furthermore, \citet{wang2024word} proposed Word-Sequence Entropy (WSE) to adjust uncertainty proportions at both the word and sequence levels based on semantic relevance, ensuring that uncertainty is aligned with the semantic importance of words within a response. 
%Word-Sequence Entropy: Towards Uncertainty Estimation in Free-Form Medical Question Answering Applications and Beyond
In addition to measuring the similarity between generated responses, ~\citet{wang2024word} proposed to judge the similarity between the target response and the generations.
%Semantic Density: Uncertainty Quantification in Semantic Space for Large Language Models
\citet{duan2023shifting} proposed Shifting Attention to Relevance (SAR), which focus on relevant components and assigns significance weights to tokens based on their contributions to the overall response.
% \citet{duan2023shifting} proposed Shifting Attention to Relevance (SAR) to enhance predictive uncertainty quantification by focusing on relevant components at both token and sentence levels. 
%Shifting attention to relevance: Towards the uncertainty estimation of large language models
%MARS: Meaning-Aware Response Scoring for Uncertainty Estimation in Generative LLMs，bakman2024mars
Unlike these carefully designed methods, \citet{yaldiz2024not} introduced a Learnable Response Scoring Function (LARS), which utilizes supervised data to capture complex token-probability dependencies. 
%Do Not Design, Learn: A Trainable Scoring Function for Uncertainty Estimation in Generative LLMs
While effective, the above methods are computationally expensive. To alleviate these computational cost, \citet{kossen2024semantic} proposed Semantic Entropy Probes (SEPs) to approximate semantic entropy by leveraging hidden states from a single generation. 
%Semantic Entropy Probes: Robust and Cheap Hallucination Detection in LLMs
% Additionlcy, by combining Verbalized Uncertainty and Probing Uncertainty, \citet{tanneru2024quantifying} and \citet{liu2024examiningllmsuncertaintyexpression} both provide novel metrics for measuring model confidence which leverage direct model prompting and perturbation analysis, offering additional insights into the uncertainty associated with natural language explanations.
%Quantifying Uncertainty in Natural Language Explanations of Large Language Models
%Examining LLMs' Uncertainty Expression Towards Questions Outside Parametric Knowledge

\textbf{Verbal confidence methods}
Due to LLMs' strong language abilities and adherence to instructions, Verbal confidence methods are proposed. For instance, one may attach the question with a prompt like ``Please respond and provide your confidence score ranging from 0 to 100.''.
% directly asking Large Language Models (LLMs) to express their uncertainty. 
\citet{xiong2023can} constructed a prompting, sampling, and aggregation framework to systematically evaluate various strategies and their integration, enabling LLMs to express their confidence in response. 
%Can LLMs Express Their Uncertainty? An Empirical Evaluation of Confidence Elicitation in LLMs.
\citet{groot2024overconfidence} proposed FaR prompting strategy, which improves the confidence calibration of LLMs by separating the fact retrieval and reflective reasoning steps.
% \citet{groot2024overconfidence}  introduced FaR prompting, enhancing calibration by separating fact retrieval and reasoning. 
%Overconfidence is Key: Verbalized Uncertainty Evaluation in Large Language and Vision-Language Models
However, verbal confidence methods face significant challenges with over-confidence.
\citet{ni2024large} found that LLMs cannot convey their uncertainties faithfully in natural language.
\citet{becker2024cycles} found that combining language confidence and proxy model probability estimation can improve the estimation of uncertainty.
%Cycles of Thought: Measuring LLM Confidence through Stable Explanations%Are Lage LanguaÅge Models More Honest in Their Probabilistic or Verbalized Confidence?
%\citet{madhusudhan2024llms} found that the language perception of LLMs is often not as accurate as the probability perception after adjusting the confidence threshold in the dataset in a determined domain.
\citet{madhusudhan2024llms} noted LLMs' language perception accuracy often lags behind probability perception, especially in specific domains
%Do LLMs Know When to NOT Answer? Investigating Abstention Abilities of Large Language Models
Furthermore, \citet{tao2024trust} found that LLMs often exhibit a high degree of overconfidence when expressing their own confidence by comparing language-based methods, consistency-based methods, and their hybrid benchmark testing methods. Their research indicates that some prompt strategies can improve the calibration of verbal confidence. 
% In most cases, consistency-based methods are superior in accuracy to language confidence, especially in arithmetic reasoning tasks, where the improvement effect is particularly significant.
%When to Trust LLMs: Aligning Confidence with Response Quality

\textbf{Internal state-based method}
Internal state-based methods suggest that the activation of the target model can be analyzed to predict the model errors.
% \citet{azaria2023internal} and \citet{liu2024uncertainty} propose techniques for assessing and predicting the reliability of LLM outputs using internal state activations. 
\citet{azaria2023internal} proposed SAPLMA by training a classifier on the hidden layer activations of an LLM to assess statement truthfulness. Similarly, \citet{liu2024uncertainty} also introduced a supervised method by  training a model on labeled datasets that analyze hidden layer activations and probability-related features.
%The internal state of an LLM knows when it’s lying. 
%Uncertainty Estimation and Quantification for LLMs: A Simple Supervised Approach
% \citet{kadavath2022language} and \citet{ji2024llm} both 
Focusing on the self-assessment capabilities of LLMs, \citet{kadavath2022language} trained models to explore the LLMs' ability to evaluate the accuracy of their responses through calibration on multiple-choice and true/false questions. 
% Complementing this, 
\citet{ji2024llm} employed a probing estimator to analyze the internal mechanisms of LLMs across various NLG tasks, assessing uncertainty before response generation.
% Language models (mostly) know what they know
%LLM Internal States Reveal Hallucination Risk Faced With a Query
Additionally, some works introduced novel interventions to refine the uncertainty estimation performance during inference. 
\citet{han2024enhancing} proposed to learn from past experience (LePe) method by leveraging historical performance records and fine-tuning instructions. 
\citet{li2024inference} presented Inference-Time Intervention (ITI) to adjust model activations selectively during inference across a limited number of attention heads, guided by a predefined set of directions. \citet{ulmer2024calibratinglargelanguagemodels} proposes a method to set confidence targets and train an additional model that predicts an LLM’s confidence based on its textual input and output.
%Enhancing Confidence Expression in Large Language Models Through Learning from Past Experience
%Inference-time intervention: Eliciting truthful answers from a language model

\textbf{Consistency-based method}
The consistency-based method is to evaluate the uncertainty of the large model through multiple generated answers.
Recently, \citet{li2024think} employed UQ sampling with perturbation and an aggregation module to quantify sampling uncertainty in text generation tasks.
%Think twice before assure: Confidence estimation for large language models through reflection on multiple answers。Li et al. (2024b) used perturbation and aggregation to measure sampling uncertainty in text generation 
\citet{pedapati2024large} proposed a paradigm to reduce overconfidence in incorrect answers by having LLMs reflect on and justify each candidate answer, then aggregating these justifications to calibrate confidence estimates.
% \citet{pedapati2024large} suggested reducing overconfidence by having LLMs justify answers and aggregate these to adjust confidence.
%Large Language Model Confidence Estimation via Black-Box Access
%\citet{becker2024cycles} proposed obtaining semantic diversity and syntactic similarity in responses via prompt perturbations, converting these into features, and then training a model to estimate confidence.
%Cycles of Thought: Measuring LLM Confidence through Stable Explanations
\citet{becker2024cycles} proposed extracting semantic diversity and syntactic similarity from perturbed prompts, training a model on these features to estimate confidence.
\citet{yang2024just} explored the stability of explanations generated by LLMs to estimate the model's confidence in its answers. 
%Just rephrase it! Uncertainty estimation in closed-source language models via multiple rephrased queries
%\citet{lin2023generating} discussed assessing confidence by combining observed consistency and self-reflection certainty to quantify uncertainty in any language model's answers.
\citet{lin2023generating} discussed combining observed consistency and self-reflection to assess language model uncertainty. 
%Generating with Confidence: Uncertainty Quantification for Black-box Large Language Models 

\paragraph{Supervised method} 
Notably, this is not an independent classification. For the convenience of comparison, the four categories above that use supervised methods are also summarized here. 

Since our research follows a supervised learning approach, we provide a complementary summary of existing supervised methods. Compared with these supervised methods, our work has fundamentally different setup, scope and application timing.

% while highlighting key differences between our approach and their
% \citet{shen2024thermometeruniversalcalibrationlarge} focused on multiple-choice QA tasks, employing an auxiliary model to calibrate token-level confidence via temperature adjustment before inference. \citet{chaudhry2024finetuninglanguagemodelsemit} fine-tuned LLMs for calibrated verbal uncertainty expressions using curated data, representing a novel approach to directly address uncertainty in a verbal way. This is not an independent classification. For the convenience of comparison, the four categories above that use supervised methods are also summarized here.
First, \citet{shen2024thermometeruniversalcalibrationlarge, chaudhry2024finetuninglanguagemodelsemit, kapoor-etal-2024-calibration} primarily focus on specific aspects of uncertainty estimation, such as classification tasks or confidence calibration in verbal or probabilistic forms, while our method aims to enhance overall performance rather than addressing isolated uncertainty measures.
%Thermometer: Towards Universal Calibration for Large Language Models.
%Finetuning language models to emit linguistic expressions of uncertainty.
%Calibration-Tuning: Teaching Large Language Models to Know What They Don’t Know.
Second, \citet{azaria2023internal, liu2024uncertainty, kadavath2022language} rely on accessible internal states for prediction, providing insights into the LLM explainary but limiting their applicability to black-box models. 
%The Internal State of an LLM Knows When It's Lying.
%Uncertainty Estimation and Quantification for LLMs: A Simple Supervised Approach
%Language Models (Mostly) Know What They Know
In comparison, our Corrector is agnostic to both response content and probability distributions, enabling broader adaptability across diverse settings. Third, \citet{ji2024llm, tao2024trust} involve more fine-grained and complex data creation processes, such as probabilistic alignment and other intricate algorithms. 
%LLM Internal States Reveal Hallucination Risk Faced With a Query
%When to Trust LLMs: Aligning Confidence with Response Quality
In contrast, our method employs a straightforward data creation and training procedure, enabling broad applicability.

\subsection{Background and Theory}
In this section, we commence by clarifying the two scales of uncertainty: \textit{relative uncertainty} and \textit{absolute uncertainty}. 
We then formalize the relative uncertainty estimation as a classification task to determine whether the target model can correctly respond to a given question. 
Subsequently, we delve into the theoretical foundations of widely-used logit-based uncertainty estimation methods, and critically examine the inherent limitations shared by those approaches that rely exclusively on target model outputs.

% Finally, we provide a theoretical justification for the advantages of our proposed method in mitigating over-confidence and underconfidence in the uncertainty estimation through bidirectional calibration.

\subsubsection{Relative Uncertainty and Absolute Uncertainty}
\label{s:relative}

Research on uncertainty estimation has led to two key concepts~\citep{kamath2020selective, vazhentsev2023hybrid}: \textit{relative uncertainty} and \textit{absolute uncertainty}, each providing distinct methods for assessing and interpreting levels of uncertainty.
Given an input $x$, a ground truth answer \(y\), and the predictive distribution of \(Y\), the predictive uncertainty for the target model regarding the input $x$ is denoted as $\text{UE}(x, \theta)$.
Relative uncertainty scores emphasize the accuracy of sample ranking, especially in discerning questions that the target model can correctly respond to from those it struggles with. Ideally, for every pair $(x_i, y_i)$ and $(x_j, y_j)$ with their predictive distributions $Y_i$ and $Y_j$, we should have
\begin{equation}    
\begin{split}
\text{UE}(x_i, \theta) &\leq \text{UE}(x_j, \theta) 
\iff \\ P(Y_i = y_i | x_i,\theta) &\geq P(Y_j = y_j | x_j,\theta).
\end{split}
\end{equation}

Stricter than relative uncertainty scores, absolute uncertainty scores support to represent the model's accuracy. 
In cases where there is an 80\% uncertainty prediction, it implies that the question is expected to be answered correctly only 20\% of the time under similar conditions.
This relationship can be mathematically expressed as
\begin{equation} 
P(Y = y | \text{UE}(x, \theta) = q) = 1 - q. 
\end{equation}

% When a model’s predicted uncertainty scores consistently align with this principle, the model is considered well-calibrated.

% Generative models have made significant strides in addressing the challenge of uncertainty in predicting responses to given questions. These methods can be broadly classified as verbal confidence, logit-based methods, and internal state-based methods.
% Generative models have made significant progress in addressing the uncertainty inherent in predicting responses to specific queries. 

% Uncertainty estimation plays a critical role across various model architectures. We specifically focus on auto-regressive models. In these models

% Existing uncertainty estimation methods for LLMs can be broadly categorized into logit-based methods, verbal confidence methods, and internal state-based methods. Due to the absence of formal theoretical frameworks for the latter two categories, we will first provide a comprehensive discussion of logit-based methods, followed by an exploration of the other two approaches.

As relative uncertainty concerns solely with the relative rankings of $h(x)=\text{UE}(x,\theta)$, it can be framed as a classification problem aimed at finding a function $h$ that minimizes the expected loss of misclassification~\citep{allikivi2024cautious, tao2023dual}. 
Consider two class labels, $\mathcal{C} = \{c_0, c_1\}$, indicating whether the targrt model can correctly answer the question or not, respectively. 
This leads to the formulation of a decision rule
\begin{equation}
g(h; \tau) = 
\begin{cases} 
c_0 & \text{if } h(x) \leq \tau \text{ (confident)} \\ 
c_1 & \text{if } h(x) > \tau \text{ (uncertain)} 
\end{cases},
\end{equation}
where $h(x)$ is a scalar measure of uncertainty and $\tau$ is the threshold. 
% The above decision rule can be simplified as $g: \mathbb{R}^n \mapsto \mathcal{C}$.
% In practice, we reject if the model is uncertain which show that the answer probably incorrect. 

Drawing from decision theory, we derive the expected loss as \textit{conditional risk} for the sample $x$:
% \begin{equation}
% \mathcal{L}_{\mathrm{UE}}(f(\boldsymbol{x}), y) = 
% \begin{cases}
% 0, & \text{if } f(\boldsymbol{x}) = y \quad \text{(correct)} \\ 
% 1, & \text{if } f(\boldsymbol{x}) \neq y \quad \text{(uncorrect)}
% \end{cases}
% \end{equation}
%  In the case of classification aimed at identifying a function $f$ that minimizes the expected loss $R(f)$ of misclassification (commonly referred to as the "risk" of $f$ in decision theory), thereby approaching the theoretical minimum known as the Bayes risk $R^*$.
\begin{equation}
% \mathrm{Risk}(x)=\lambda_{i,1-i}P(c_{1-i}\mid x)
\mathrm{Risk}(x)=\lambda_{c_i,c_{1-i}}h_{c_{1-i}}(x),
\label{eq:risk}
\end{equation}
where $c_i,  i \in \{0, 1\}$ denotes the true label of the sample $x$, and $h_{c_{1-i}}(x) = P(c_{1-i} \mid x)$ is the posterior probability of misclassifying the sample $x$ as class $c_{1-i}$. $\lambda_{c_i,c_{1-i}}$ represents the loss associated with this misclassification—specifically, a penalty incurred when the sample with the label $c_i$ is classified as $c_{1-i}$. 
% In this context, the parameter $\lambda$ quantifies the cost of different types of classification errors, allowing the model to adjust its decisions according to the severity of those errors.
Our task is to find $h^*$ that minimizes the overall risk
\begin{equation}
\mathrm{Risk}(h)=\mathbb{E}_x\left[\mathrm{Risk}(h(x))\mid x\right].
\label{eq:rh}
\end{equation}

% \begin{equation}
% \mathrm{Risk}(f,g;\tau)=\frac{\mathbb{E}_{p_\text{data}(x,y)}[g(x;\tau)\mathcal{L}_\text{SC}(f(x),y)]}{\mathbb{E}_{p_\text{data}(x,y)}[g(x;\tau)]} ,
% \end{equation}

\subsubsection{Theoretical Foundations of Uncertainty Estimation for LLM}
\label{s:tfue}
LLMs typically generate outputs in an auto-regressive manner, which iteratively predict the probability distribution of the subsequent token based on the evolving context~\citep{gregor2014deep}. 
Given an input sequence $x$ with the objective of generating an output sequence $y = \{y_1, y_2, \ldots, y_L\}$, the conditional probability of the $l$-th token $y_l$ is denoted as $P(y_l | y_{<l}, x; \theta)$. 
This probability depends on all previously generated tokens $y_{<l} = \{y_1, y_2, \ldots, y_{l-1}\}$ as well as the input $x$. 
The probability of generating the entire sequence $y$ can be expressed as the product of the conditional probabilities of each individual token:
\begin{equation}
P(y | x; \theta) = \prod_{l=1}^L P(y_l | y_{<l}, x; \theta),
\end{equation}
where 
$P(y_l|y_{<l},x;\theta)=\frac{e^{z_l/T}}{\sum_je^{z_j/T}}$,
$z$ is the raw logit, and $T$ is the temperature that controls the smoothness of the probability distribution. 
This posterior probability provides a probabilistic framework for sequence generation. 
Moreover, according to prior research~\citep{malinin2020uncertainty}, the total uncertainty for the generation of $y$ is given by the entropy of the predictive posterior:
\begin{equation}
\begin{aligned}
\text{PE}(x) &= \mathcal{H}[P(y\mid x,\theta)] \\
    &= \mathbb{E}_{P(y\mid x,\theta)}[-\ln{P}(y\mid x,\theta)] \\
    &= -\sum_{y\in Y}P(y\mid x,\theta)\ln P(y\mid x,\theta).
\label{eq:pe}
\end{aligned}
\end{equation}

% Malinin and Gales (2021)
% combine \ref{eq:pe} and \ref{eq:rh}, we get:
% \begin{equation}
% R(f)=\lambda_{i,1-i}\left[\delta_{i,0}\mathbb{E}_{P(y\mid x,\theta)}[-\ln P(y\mid x,\theta)]+\delta_{i,1}\left(1-\mathbb{E}_{P(y\mid x,\theta)}[-\ln P(y\mid x,\theta)]\right)\right]
% \end{equation}
% $\delta_{i,0}$ and $\delta_{i,1}$ are indicator functions.

In practice, due to the exponential computational complexity of traversing the entire response set, Monte Carlo approximation method~\citep{papadopoulos2001uncertainty} is employed via beam search with a single target model for generation. The approximate entropy is defined as
\begin{equation}
    PE(x) \approx - \frac{1}{B} \sum_{b=1}^{B} \ln P(y_b|x, \theta),
\end{equation}
where $P(y_b|x, \theta)$ denotes the posterior probability of the $b$-th beam search candidate.
% Kuhn et al. (2023) 
Base on these, \citet{kuhn2023semantic} proposed to cluster generations with similar meanings and compute entropy using the probabilities associated with each semantic cluster. 
This approach is formulated as
\begin{equation}
SE(x, \theta) = -\frac{1}{C} \sum_{i=1}^{C} \ln P(c_i | x, \theta),
\end{equation}
where $c_i$ denotes each semantic cluster and $C$ represents the set of all clusters. 

Another form of improvement is to assign weights to each token in the generation when calculating posterior probabilities ~\citep{duan2023shifting, bakman2024mars}, either through a manually designed algorithm or a training way, which can be formulated as
\begin{equation}
\tilde{P}(y\mid x; \theta) = \prod_{l=1}^{L} P(y_l \mid y_{<l}, x; \theta) \cdot w_l,
\end{equation}
where $w_l$ represents the weight assigned to the $l$-th token.

\subsection{Compare with Other Supervised UE Method }
\label{app:comparative}

\begin{table*}[tb]
\centering
\resizebox{\linewidth}{!}{
\begin{tabular}{lccc|ccc|ccc}
\toprule
\textbf{Method} & \multicolumn{3}{c}{\textbf{AUROC ($\uparrow$)}} & \multicolumn{3}{c}{\textbf{ECE ($\downarrow$)}} & \multicolumn{3}{c}{\textbf{F1 ($\uparrow$)}} \\
\cmidrule{2-4} \cmidrule{5-7} \cmidrule{8-10}
 & \textbf{Vanilla} & \textbf{+Corrector} & \textbf{+Wb-S} & \textbf{Vanilla} & \textbf{+Corrector} & \textbf{+Wb-S} & \textbf{Vanilla} & \textbf{+Corrector} & \textbf{+Wb-S} \\
\midrule

\textbf{PE} & 64.52  & \textbf{69.76}  & \textbf{64.37}   & 21.38  & \textbf{17.24}  & \textbf{20.18}   & 47.54  & \textbf{64.97}  & 52.45 \\

\textbf{LN-PE} & 72.55  & \textbf{74.79}  & \textbf{72.75}  & 14.31  & \textbf{11.53}  & \textbf{14.31}  & 52.83  & \textbf{61.99}  & \textbf{52.83} \\

\textbf{SE} & 80.92  & \textbf{82.12}  & \textbf{80.92}  & 13.07  & \textbf{12.76}  & \textbf{13.07}  & 57.14  & \textbf{65.75}  & \textbf{57.14} \\

\textbf{SAR-t} & 79.55  & \textbf{79.93}  & \textbf{79.55}  & 16.40  & \textbf{13.70}  & \textbf{16.40}  & 51.67  & \textbf{57.60}  & \textbf{51.67} \\

\textbf{SAR-s} & 69.87  & \textbf{77.09}  & \textbf{69.87}  & 23.17  & \textbf{20.00}  & \textbf{23.17}  & 53.44  & \textbf{66.22}  & \textbf{53.44} \\

\textbf{SAR} & 80.92  & \textbf{81.90}  & \textbf{80.92}  & 16.17  & \textbf{13.76}  & \textbf{16.17}  & 51.20  & \textbf{59.54}  & \textbf{51.20} \\

\bottomrule
\end{tabular}
}
\caption{Comparison of performance between the \textit{Corrector} and the Wb-S method.}
\label{tab:comparison}
\end{table*}

Following publicly available code and experimental settings, we compare our \textit{Corrector}'s performance with the supervised UE method provided by \citet{liu2024uncertainty}, which we refer to as Wb-S.
We focused on enhancement after applying the supervised method to the strong unsupervised baselines, targeting model LLaMA-3-8B-Instruct on dataset TriviaQA. 

% other supervised learning methods to verify the effectiveness of our proposed Corrector in enhancing uncertainty estimation. Limited by the availability of code and experimental settings, only the supervised method proposed by \citet{liu2024uncertainty} was selected for comparative evaluation. 

As shown in Table~\ref{tab:comparison}, our \textit{Corrector} achieved harmonized  improvements on all strong baselines. However, when we replaced \textit{Corrector} with the Wb-S method, integrated into our pipeline, only marginal improvements were observed with the Predictive Entropy (PE) method, and no significant effects were noted with other unsupervised methods, and thus the metrics are almost indistinguishable from vanilla's after using the Wb-S method.
% Our evaluation method yields more accurate AUROC scores due to the introduction of more rigorous and scientific criteria for assessing response correctness, combining rule-based and LLM-based approaches. This may partly explain the discrepancy between our replication results and those presented in the original paper. 

% However, the fairness of our evaluation approach is confirmed by the consistent performance of other established strong unsupervised baselines (e.g., SE, SAR) under our evaluation, compared to their published results. Additionally, compare to the requested supervised baseline, we achieved more comprehensive improvements beyond AUROC, and our method's applicability extends beyond white-box scenarios. All these comparative analyses further underscored the superiority of the \textit{Corrector}. 

\subsection{Generalization}
\label{app:gene}

% In previous experiments, we train the \textit{Corrector} on dataset $\mathcal{D^*}_{\text{cor}}$, which is curated to align with the capabilities of a target model within a particular domain of knowledge.
The previous results indicate the \textit{Corrector}'s effectiveness on the in-distribution evaluation set. In the subsequent analysis, we investigate its \textbf{cross-dataset and cross-model generalization} capabilities.

% highlight two out-of-distribution scenarios: \textbf{domain of data}, \textbf{target model}.

% However, our analysis reveals two primary variables that can lead to out-of-distribution scenarios: the data type and the target model. Accordingly, we will assess the Corrector's generalization capabilities on additional out-of-distribution datasets and models.
% Consequently, we will evaluate the Corrector's generalization capabilities on additional out-of-distribution datasets and models.

\paragraph{Cross-Dataset Generalization}
% we investigate the \textbf{Corrector} performance when the target model encounters questions that deviate from the training data distribution.

In terms of cross-dataset (different domain) generalization, we tested the Corrector's generalization via cross-training (training on TriviaQA and then testing on SciQA, and training on SciQA and then testing on TriviaQA, with OPT-2.7B as the target model). These datasets differ significantly in question domain (trivia vs. scientific) and size (10,000 vs. 1,000 examples), presenting a significant generalization challenge. We observed average absolute improvements of 4\%-6\% over baselines.

To further investigate cross-dataset (same domain) generalization, we conducted additional experiments using \textbf{MedMCQA} ~\cite{pal2022medmcqa} and \textbf{MedQA}~\cite{jin2020diseasedoespatienthave} (distinct but related medical datasets). Using the same setup as different domain, we observed absolute improvements of 7-11\%. This demonstrates promising generalization within the same domain. The results are shown in the Table~\ref{(a)}.

%To evaluate the generalization capability of our \textit{Corrector} across different data domains, we conduct experiments by training the \textit{Corrector} on the dataset $\mathcal{D^*}_{\text{cor}}$, crafted from either TriviaQA or SciQA, and then evaluating it on the alternate one.
%As illustrated in Table \ref{tab:gene}, the \textit{Corrector} achieves optimal performance when both training and evaluating occur within the same data domain. 
%Remarkably, even when training and evaluating on different domains, the \textit{Corrector} still demonstrates a enhancement, yielding an average improvement of approximately 0.05.
%One possibility is that the target model exhibits comparable knowledge proficiency across both data domains.

\paragraph{Cross-Model Generalization}
In terms of cross-model generalization, we tested the Corrector's generalization across models (OPT-2.7B, OPT-6.7B, LLaMA-3-8B) on SciQA, reporting average AUROC improvement over all baselines. 

Experimental results show the \textit{Corrector} generalizes well within the same model family (e.g., OPT-2.7B and OPT-6.7B, with average absolute improvements of 6\%-11\%), likely due to their similar performance on the original SciQA dataset. We observed limited transferability across LLMs with significantly different performance and architectures (e.g., LLaMA-3-8B and OPT-2.7B). However, even in these scenarios, CUE achieves an average absolute improvement of 3\%, which demonstrates the efficacy of our method.

%We investigate the generalization for target model by training \textit{Corrector} on \(\mathcal{D^*}_{\text{cor}}\) sourced from a different target model than the one used for evaluating.
%As shown in Table~\ref{tab:gene}, in cases where models exhibit relatively comparable knowledge capabilities, such as OPT-2.7B and OPT-6.7B, the \textit{Correcter} exhibits generalization ability, yielding average AUROC improvements of 0.11 and 0.06, respectively.
%Conversely, when a substantial performance gap exists between models, such as OPT-2.7B and LLaMA-3-8B-Instruct, 
% while we achieve an average AUROC improvement of 0.03, we contend that this does not represent true generalization, especially since some baselines from RBS, perform even worse than random guessing.
%we achieve an average AUROC improvement of 0.03. When focusing solely on the challenging baselines from CBS, the improvement drops to 0.01.

\begin{table*}[t]
    \centering
    \vspace{-10mm}
        \resizebox{\linewidth}{!}{
        \subfloat[][Generalization for Domain of Data]{
            \centering
            \begin{minipage}[t]{0.4\textwidth}
            \centering

            \begin{tabular}{lcc}
            % \label{tab:g_data}
                \toprule
                \multicolumn{1}{c}{} & \multicolumn{1}{c}{TriviaQA} & \multicolumn{1}{c}{SciQA} \\
                \midrule
                TriviaQA & 19.59 & 4.05 \\
                SciQA & 6.03 & 10.20 \\
                \bottomrule
            \end{tabular} 
            \end{minipage}
            %\hfil

            \begin{minipage}[t]{0.4\textwidth}
            \centering
            \begin{tabular}{lcc}
            % \label{tab:g_data}
                \toprule
                \multicolumn{1}{c}{} & \multicolumn{1}{c}{MedMCQA} & \multicolumn{1}{c}{MedQA} \\
                \midrule
                MedMCQA & 15.54 & 10.70 \\
                MedQA & 6.93 & 11.21 \\
                \bottomrule
            \end{tabular}
            \end{minipage}
            
            \label{(a)}
        } \quad 
        \subfloat[][Generalization for Target Model]{
            \centering
            % \begin{tabular}{lcc}
            %     \toprule
            %     \multicolumn{1}{c}{} & \multicolumn{1}{c}{OPT-2.7B} & \multicolumn{1}{c}{LLaMA3-8B} \\
            %     \midrule
            %     OPT-2.7B & 19.59 & 3.43 \\
            %     LLaMA3-8B & 3.23 & 11.21 \\
            %     \bottomrule
            % \end{tabular}
            \begin{tabular}{lccl}
                \toprule
                \multicolumn{1}{c}{} & \multicolumn{1}{c}{OPT-2.7B} & \multicolumn{1}{c}{OPT-6.7B} & \multicolumn{1}{c}{LLaMA3-8B}\\
                \midrule
                OPT-2.7B & 19.59 & 11.80  & 3.23\\
                OPT-6.7B & 6.08 & 11.21  & 3.43\\
                \bottomrule
            \end{tabular}
            \label{(b)}
        }} \\
        
        %     \subfloat[][Generalization for Task]{
        %         \resizebox{\linewidth}{!}{
        %         \centering
        %         \begin{tabular}{lccc|ccc|ccc}
        %             \toprule
        %             \textbf{Method} & \multicolumn{3}{c}{\textbf{AUROC $\uparrow$}} & \multicolumn{3}{c}{\textbf{ECE $\downarrow$}} & \multicolumn{3}{c}{\textbf{F1 $\uparrow$}} \\
        %             \cmidrule(r){2-4} \cmidrule(r){5-7} \cmidrule(r){8-10}
        %             & \textbf{Vanilla} & \textbf{+Corr} & \textbf{Improv} & \textbf{Vanilla} & \textbf{+Corr} & \textbf{Improv} & \textbf{Vanilla} & \textbf{+Corr} & \textbf{Improv} \\
        %             \midrule
        %             SAR & 60.16 & 66.63 & +6.47 & 52.11 & 23.76 & -28.35 & 13.98 & 39.85 & +25.87 \\
        %             SAR-s & 54.23 & 60.10 & +5.87 & 58.13 & 27.88 & -30.25 & 62.25 & 78.54 & +16.29 \\
        %             SAR-t & 60.35 & 66.49 & +6.14 & 56.84 & 26.44 & -30.40 & 11.04 & 32.13 & +21.09 \\
        %             SE & 58.24 & 63.71 & +5.47 & 62.19 & 33.18 & -29.01 & 46.49 & 55.81 & +9.32 \\
        %             LN-PE & 62.96 & 69.11 & +6.15 & 52.79 & 25.44 & -27.35 & 26.32 & 44.59 & +18.27 \\
        %             PE & 55.30 & 61.42 & +6.12 & 58.11 & 23.82 & -34.29 & 7.54 & 39.73 & +22.19 \\
        %             LS & 47.92 & 53.27 & +5.35 & 46.86 & 16.35 & -30.51 & 6.96 & 33.51 & +26.55 \\
        %             \bottomrule
        %         \end{tabular}
        %         \label{(c)}
        %     }
        % }\\

    \caption{Average AUROC scores (\%) improvement of after appling our method to baselines. (a) The leftmost column indicates the domains of data used in training, while the topmost row represents the domains of data used for evaluating, with OPT-2.7B serving as the target model. (b) The leftmost column denotes the target model during training, whereas the topmost row signifies the target model during evaluating, with TriviaQA utilized as the target domain of data.}
    \label{tab:gene}
    \vspace{-1mm}
\end{table*}

% \paragraph{Task}
% For fairness and representativeness, our main experiments were set up on the same datasets as previous works on QA tasks, which were representative of LLM functions and a common focus of uncertainty estimation research. This allowed us to directly demonstrate the improvement of our method compared to the current UE technology.

% Using the same setup as described above, we evaluated our corrector against established baselines, focusing on coordinated improvements in indication (AUROC), balance (F1 score), and calibration (ECE), using OPT-6.7B as the target model.

\subsection{Statistical Hypothesis Testing} 
\label{t-test}

Regarding the performance improvements compared to other robust UE methods, our approach provided harmonized, multi-dimensional enhancements across various aspects of uncertainty estimation, including \textit{indication}, \textit{balance}, and \textit{calibration}. To statistically validate the significance of these improvements across all metrics, we conducted \textbf{t-tests} on the TriviaQA and SciQA datasets, comparing our method against strong baselines (SE, t-SAR, s-SAR, SAR). The results yielded \textbf{p < 0.05} for each baseline on both datasets, demonstrating that the performance improvements were statistically significant. 

It was also essential to clarify that the reported performance of the SE and SAR methods reflected their saturation point. This indicated that further increasing the number of samples—commonly used to enhance their performance—no longer resulted in additional gains. In contrast, our method surpassed this saturation point, effectively addressing the limitations of these methods and delivering continued improvements.

\begin{table*}[htbp]
    \centering
    \resizebox{\textwidth}{!}{%
        \begin{tabular}{lccc|ccc|ccc|ccc}
            \toprule
            \multirow{2}{*}{Method} & \multicolumn{3}{c}{TriviaQA AUROC$\uparrow$} & \multicolumn{3}{c}{TriviaQA ECE $\downarrow$} & \multicolumn{3}{c}{SciQA AUROC$\uparrow$} & \multicolumn{3}{c}{SciQA ECE $\downarrow$} \\
            \cmidrule(lr){2-4} \cmidrule(lr){5-7} \cmidrule(lr){8-10} \cmidrule(lr){11-13}
            & Vanilla & w/ Corrector & Improv & Vanilla & w/ Corrector & Improv & Vanilla & w/ Corrector & Improv & Vanilla & w/ Corrector & Improv \\
            \midrule
            Correcter & 69.87 & - & - & 6.73 & - & - & 65.38 & - & - & 18.19 & - & - \\
            LS & 19.57 & 69.82 & 50.25 & 70.25 & 7.41 & -62.84 & 53.67 & 65.38 & 11.71 & 38.64 & 18.19 & -20.45 \\
            VC & 62.34 & 74.89 & 12.55 & 23.41 & 16.78 & -6.63 & 68.22 & 72.15 & 3.93 & 31.88 & 19.47 & -12.36 \\
            P(True) & 57.14 & 72.29 & 15.15 & 24.67 & 19.84 & -4.83 & 65.63 & 71.41 & 5.78 & 34.56 & 31.92 & -2.64 \\
            PE & 64.52 & 69.76 & 5.25 & 21.38 & 17.24 & -4.13 & 66.54 & 67.98 & 1.44 & 40.67 & 34.07 & -6.60 \\
            LN-PE & 72.55 & 74.79 & 2.24 & 14.31 & 11.53 & -2.79 & 69.48 & 71.56 & 2.08 & 29.38 & 23.76 & -5.62 \\
            SE & 80.92 & 82.12 & 1.20 & 13.07 & 12.76 & -0.31 & 71.59 & 72.93 & 1.34 & 30.54 & 25.23 & -5.31 \\
            SAR-t & 79.55 & 79.93 & 0.38 & 16.40 & 13.70 & -2.70 & 72.26 & 73.87 & 1.61 & 30.37 & 26.81 & -3.56 \\
            SAR-s & 69.87 & 77.09 & 7.22 & 23.17 & 20.00 & -3.17 & 74.96 & 75.72 & 0.76 & 38.54 & 36.18 & -2.37 \\
            SAR & 80.92 & 81.90 & 0.98 & 16.17 & 13.76 & -2.41 & 73.88 & 75.19 & 1.31 & 28.97 & 25.60 & -3.37 \\
            \bottomrule
        \end{tabular}
    }
    \caption{Comparison of Pure Corrector Performance with the Baseline, and the Performance Gains from Using Corrector Scores as Corrections to the Baseline.}
    \label{tab:pure_performance}
\end{table*}

% Pure Corrector Performance与baseline的对比，以及Corrector的分数作为baseline的修正分数带来的性能提升。

\subsection{Detail of Hyperparameter w*}
\label{app:detail_of_w}
\subsubsection{Sensitivity of Hyperparameter w*}
To address test the sensitivity of the hyperparameter $w^*$, we conducted a sensitivity analysis by performing tests on the opt-6.7b and llama3-8b models using the TriviaQA dataset. We adjusted the $w^*$ values around the original optimal value in thousandth-increments and recorded the AUROC performance under different $w^*$ values. When the performance change was within 1\%, we recorded this range as the ``stable range''.
Table \label{tab:sensitivity_of_hyperparameter} illustrates the $w^*$ stable ranges for various UE methods. We observed that, in the majority of cases, performance fluctuations remained below 1\% within a $w^*$ range of 0.107 to 0.442, indicating a degree of robustness in the model to $w^*$.

\begin{table}[htbp]
    \centering
    \resizebox{0.5\textwidth}{!}{%
        \begin{tabular}{lcc|cc}
            \toprule
            Method+Cor & \multicolumn{2}{c}{OPT-6.7B} & \multicolumn{2}{c}{LLaMA3-8B} \\
            \cmidrule(r){2-3} \cmidrule(r){4-5}
            & $w^*$ Range & Range Difference & $w^*$ Range & Range Difference \\
            \midrule
            SAR & 0.12 - 0.362 & 0.242 & 0.367 - 0.809 & 0.442 \\
            SAR-s & 0.028 - 0.427 & 0.399 & 0.539 - 0.844 & 0.305 \\
            SAR-t & 0.038 - 0.318 & 0.28 & 0.447 - 0.733 & 0.286 \\
            SE & 0.246 - 0.51 & 0.264 & 0.701 - 0.808 & 0.107 \\
            LN-PE & 0.223 - 0.374 & 0.151 & 0.797 - 0.926 & 0.129 \\
            PE & 0.117 - 0.323 & 0.206 & 0.759 - 0.945 & 0.186 \\
            LS & 0.458 - 1.0 & 0.542 & 0.993 - 1.0 & 0.007 \\
            \bottomrule
        \end{tabular}
    }
    \caption{Stable Ranges of $w^*$ for Different UE Methods on TriviaQA (within 1\% AUROC change).}
    \label{tab:sensitivity_of_hyperparameter}
\end{table}

\subsubsection{Hyperparameter w* Setting in Cross-domain Experiments}
 By default, we select the optimal weight $w^*$ using the same (training) domain dev set. However, as shown in the supplementary experiments addressing R1, $w^*$ exhibits elasticity. Thus, theoretically, using $w^*$ from a truly cross-domain set would partially retain the Corrector's enhancement effect.

We validated the cross-domain experiment setup (train on SciQA/TriviaQA and test on TriviaQA/SciQA) mentioned by the reviewer, selecting $w^*$ using both in-domain and cross-domain validation sets, and recorded the resulting average AUROC improvements across methods. The experimental results are shown in Table \ref{cross_domain_performance_for_w}.

\begin{table}[htbp]
    \centering
    \resizebox{0.5\textwidth}{!}{%
        \begin{tabular}{lcc}
            \toprule
            Training => Testing Domain & Same (training) domain $w^*$ & Cross-domain $w^*$ \\
            \midrule
            SciQA => TriviaQA & 6.03 & 3.92 \\
            TriviaQA => SciQA & 4.05 & 2.75 \\
            \bottomrule
        \end{tabular}
    }
    \caption{Average AUROC improvements (\%) across methods when selecting $w^*$ using in-domain and cross-domain validation sets for cross-domain experiments.}
    \label{cross_domain_performance_for_w}
\end{table}

\subsection{Pure Corrector Performance}
\label{app:pure_corrector}
To provide more insights about pure Corrector performance, we combine and present relevant data from Table \ref{tab:main_res} and Table \ref{tab:ablation} of our paper, as shown in the table \ref{tab:pure_performance}.

On TriviaQA, the pure performance of the Corrector outperformed LS, VC, P(True), and PE, but underperformed SE, SAR-t, SAR-s, and SAR. On SciQA, the pure performance of the Corrector outperformed LS, performed similarly to VC, P(True), PE, and LN-PE, but underperformed SE, SAR-t, SAR-s, and SAR. Across all ``w/ Corrector'' settings, we saw universal improvement, showing Corrector's great complementarity with unsupervised methods. Regarding the ECE metric, we observed that the calibration of the lightweight model is significantly better than strong UE methods, which makes the combination of the two even more advantageous.

\
\end{document}